\documentclass[10pt,twocolumn,letterpaper]{article}

\usepackage[pagenumbers]{cvpr} 

\definecolor{cvprblue}{rgb}{0.21,0.49,0.74}
\usepackage[pagebackref,breaklinks,colorlinks,allcolors=cvprblue]{hyperref}
\usepackage{algorithm}
\usepackage{algorithmic}
\usepackage{bm}
\usepackage{booktabs}
\usepackage{amsmath}
\usepackage{multirow}
\usepackage{amsfonts}
\usepackage{makecell}
\usepackage{colortbl}
\definecolor{magenta-link}{RGB}{199, 21, 133}
\definecolor{c1}{RGB}{255,153,153} 
\definecolor{c2}{RGB}{255,204,153} 

\def\paperID{566} 
\def\confName{CVPR}
\def\confYear{2026}

\title{STAvatar: Soft Binding and Temporal Density Control for Monocular 3D Head Avatars Reconstruction}

\author{
	Jiankuo Zhao$^{1,2}$, Xiangyu Zhu$^{1,2}$, Zidu Wang$^{1,2}$, Zhen Lei$^{1,2,3,4}$\thanks{Corresponding author}\\
	$^{1}$MAIS, Institute of Automation, Chinese Academy of Sciences\\
	$^{2}$School of Artificial Intelligence, University of Chinese Academy of Sciences\\
	$^{3}$CAIR, HKISI, Chinese Academy of Sciences \quad $^{4}$SCSE, FIE, M.U.S.T\\
	{\tt\small \{zhaojiankuo2024, xiangyu.zhu, zhen.lei\}@ia.ac.cn}
}

\begin{document}
\maketitle
\begin{abstract}
Reconstructing high-fidelity and animatable 3D head avatars from monocular videos remains a challenging yet essential task. Existing methods based on 3D Gaussian Splatting typically bind Gaussians to mesh triangles and model deformations solely via Linear Blend Skinning, which results in rigid motion and limited expressiveness. Moreover, they lack specialized strategies to handle frequently occluded regions (e.g., mouth interiors, eyelids). To address these limitations, we propose STAvatar, which consists of two key components: (1) a UV-Adaptive Soft Binding framework that leverages both image-based and geometric priors to learn per-Gaussian feature offsets within the UV space. This UV representation supports dynamic resampling, ensuring full compatibility with Adaptive Density Control (ADC) and enhanced adaptability to shape and textural variations. (2) a Temporal ADC strategy, which first clusters structurally similar frames to facilitate more targeted computation of the densification criterion. It further introduces a novel fused perceptual error as clone criterion to jointly capture geometric and textural discrepancies, encouraging densification in regions requiring finer details. Extensive experiments on four benchmark datasets demonstrate that STAvatar achieves state-of-the-art reconstruction performance, especially in capturing fine-grained details and reconstructing frequently occluded regions. Our project is available at \href{https://jiankuozhao.github.io/STAvatar/}{\color{magenta-link}{https://jiankuozhao.github.io/STAvatar/}}.
\end{abstract}

\section{Introduction}
Reconstructing animatable and photo-realistic 3D head avatars has been a longstanding problem in computer vision and graphics \cite{egger20203d,saito2024relightable,wang20243d,GUAVA,wang2025pc,wang2025srm}, driven by growing demands in AR/VR, telepresence, digital humans and interactive media. To achieve high-quality avatar reconstruction, conventional methods typically rely on expensive multi-camera systems \cite{guo2019relightables,yang2023towards} which require complex manual setups and are impractical for consumer-level deployment. In contrast, monocular-based approaches \cite{tang2025gaf,zheng2022avatar,grassal2022neural} aim to produce high-fidelity results using a single consumer-grade camera, enabling broader accessibility in immersive and personalized virtual environments.

\begin{figure}[t]
	\centering
	\includegraphics[width=0.95\linewidth]{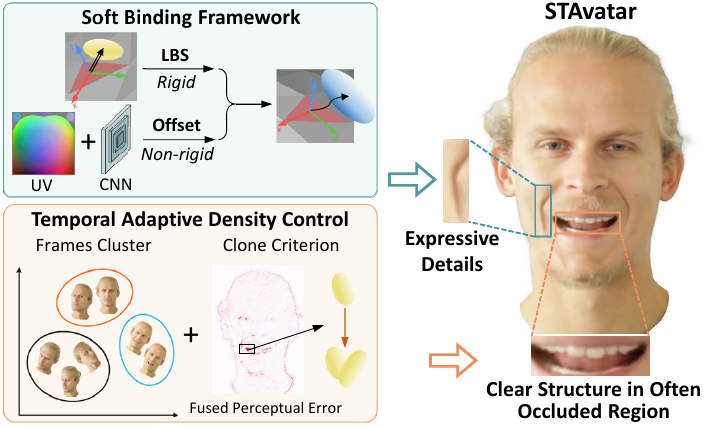}
	\caption{STAvatar proposes a Soft Binding framework and a Temporal Adaptive Density Control strategy to reconstruct high-fidelity 3D head avatars from monocular videos.}
	\vspace{-10pt}
	\label{fig:brief}
\end{figure}

Recent years have witnessed the emergence of 3D Gaussian Splatting (3DGS) \cite{kerbl3Dgaussians} as a powerful volumetric representation, delivering both high-fidelity rendering and real-time performance in neural scene reconstruction. Building on its success in static scenes, many works have explored its application to animatable 3D head avatar reconstruction. These methods typically follow a two-stage pipeline: a parametric head-tracking algorithm \cite{qian2024vhap,zielonka2022towards} is first employed to estimate the facial pose, expression, and shape for each frame; subsequently, Gaussian primitives are bound to mesh triangles and driven by a deformation field, which is typically implemented using Linear Blend Skinning (LBS) \cite{qian2023gaussianavatars}. During training, each Gaussian’s parameters are optimized via differentiable rendering to minimize photometric loss. Despite achieving enhanced visual fidelity and animation realism, existing Gaussian-based approaches still face several open challenges. To address these limitations, we propose STAvatar, as shown in Fig.~\ref{fig:brief}, and provide a detailed explanation below.

\begin{figure}[t]
	\centering
	\includegraphics[width=1\linewidth]{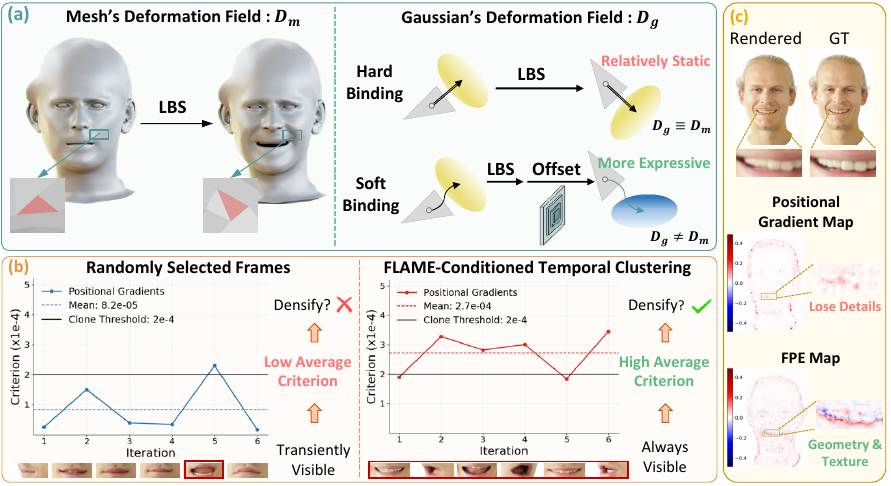}
	\caption{Limitations of existing research. (a) Hard binding forces Gaussians to remain relatively static within the triangle coordinate frames, thereby limiting their ability to capture fine-grained details. (b) Transiently visible regions, such as mouth interiors, often exhibit low average positional gradients, which impedes effective Gaussian densification. (c) The positional gradient only reflects geometric inconsistencies and often loses texture details, which hinders the addition of Gaussians in high-frequency regions.}
	\vspace{-10pt}
	\label{fig:drawback}
\end{figure}

First, existing animatable reconstruction methods often suffer from rigid connection between Gaussian primitives and mesh triangles, resulting in poor handling of non‑rigid and fine‑scale deformations. Fig.~\ref{fig:drawback}(a) illustrates this simple hard binding framework \cite{zhang2025fate,SplattingAvatar:CVPR2024, HRAvatar} in which each Gaussian deforms solely according to the LBS of its parent triangle. Although this enables basic animation control, it fails to model fine-grained deformations, as the Gaussians remain relatively static within the triangle's local coordinate frame. This limitation stems from the gap between mesh and Gaussian representations. While meshes inherently encode topology and support native skinning for deformation, Gaussians are structureless and rely on externally defined deformation fields to capture complex motions and fine details such as facial wrinkles. Some methods \cite{chen2024monogaussianavatar,xiang2024flashavatar,giebenhain2024npga} attempt to enhance each Gaussian’s deformation using a fixed-dimension multilayer perceptron (MLP). However, this design requires predefining a fixed number of Gaussians, which is suboptimal since different identities often demand varying Gaussian densities. Besides, it neglects spatial context and may be incompatible with the Adaptive Density Control (ADC) inherent to 3DGS. To enhance deformation ability while preserving the flexibility of ADC, STAvatar proposes a UV-Adaptive Soft Binding framework, which integrates the LBS for coarse deformation and a dual-branch network for fine-detail recovery. By generating a feature offset map in UV space, STAvatar leverages spatial context to model the Gaussian deformation field and enables arbitrary-resolution sampling to support ADC.

Second, the vanilla ADC in 3DGS is tailored for static scene reconstruction and fails to handle transiently  visible or frequently occluded regions in dynamic avatar reconstruction. As illustrated in Fig.~\ref{fig:drawback}(b), certain regions (\textit{e.g.}, mouth interiors) are only visible in a subset of frames and contribute minimally to rendering when occluded. This results in low average criterion-such as the positional gradient used in 3DGS-which impedes sufficient densification. In addition, as shown in Fig.~\ref{fig:drawback}(c), the positional gradient in 3DGS only captures geometric discrepancies while neglecting texture details \cite{rota2024revising}, which leads to the omission of regions with high texture errors and the failure to add Gaussians to these regions. To address these limitations of ADC, STAvatar propose a Temporal Adaptive Density Control strategy incorporating FLAME-Conditioned Temporal Clustering (FTC) and a Fused Perceptual Error with Average-Peak Criterion (FPE-AP). As shown in Fig.~\ref{fig:drawback}(b), FTC partitions frames into structure‑coherent clusters to encourage frequently occluded regions remain visible within a cluster, facilitating targeted refinement. FPE‑AP replaces positional gradients by directly estimating each Gaussian's fused perceptual error, which jointly considers both geometric and textural discrepancies. Besides, FPE-AP selects Gaussians exhibiting the highest instantaneous errors across iterations to better capture regions with large peak errors. 

Finally, we evaluate STAvatar through extensive experiments on four datasets, demonstrating clear improvements over existing methods in recovering fine details and handling challenging regions such as mouth interiors and eyelids. Besides, STAvatar exhibits accurate cross-reenactment performance, highlighting its potential in real-world applications. Our main contributions are as follows:
\begin{itemize}
    \item We propose a UV‑Adaptive Soft Binding framework that supports both soft binding and ADC of 3DGS in animatable Gaussian avatar reconstruction.
    \item We propose a Temporal ADC strategy incorporating FTC and FPE-AP which enhances the ADC in dynamic Gaussian avatar reconstruction.
    \item We conduct experiments on various datasets, demonstrating superior rendering quality, improved detail reconstruction, and more accurate cross-reenactment.
\end{itemize}
\section{Related Work}

\subsection{Gaussian Head Reconstruction}
3DGS \cite{kerbl3Dgaussians} has recently emerged as a flexible and efficient approach for 3D scene representation. By employing a large number of anisotropic Gaussian primitives—each parameterized by geometry and appearance attributes—it achieves high-quality reconstructions, making it particularly well-suited for detailed human head modeling. To enable avatar driving, typical methods \cite{qian2023gaussianavatars,liu2025lucas,tang2025gaf,saunders2025gasp,SplattingAvatar:CVPR2024, zhang2025fate} adopt a hard-binding framework that associates Gaussians with mesh triangles to transfer mesh motion via LBS, which, however, fails to bridge the inherent gap between mesh and Gaussian representations, leading to suboptimal performance in handling complex motions and expressions. Alternatively, other approaches \cite{xiang2024flashavatar,chen2024monogaussianavatar,xu2024gaussian,liao2023hhavatar} leverage MLP to predict per-Gaussian attribute offsets, thereby achieving a more flexible soft-binding framework. However, these methods often neglect ADC by using a fixed number of Gaussians across different identities, which limits expressiveness and reconstruction quality. In addition, several studies \cite{ma20243d,li2025rgbavatar,zielonka2025gaussian} propose to represent avatars with a base model and a set of expression blendshapes. Although these methods can achieve high training efficiency, they often suffer from low fidelity or produce blurred results under large pose variations, since the learned blendshapes may not adequately capture diverse pose-dependent deformations.

\subsection{Adaptive Density Control of 3DGS}
ADC strategy is a key module in 3DGS which dynamically clones, splits and prunes Gaussians to balance detail preservation and representation compactness. The vanilla ADC selects primitives for densification based on average positional gradients and Gaussian scales. However, this heuristic often results in underfitting and blurred details, as it neglects texture-related cues and only focus average criterion. To overcome these limitations, some recent works \cite{rota2024revising,zhang2024pixel,kim2024color,mallick2024taming,zhou2025perceptual} revise the densification criterion to incorporate richer information, thereby achieving finer details in underrepresented regions. Besides, optimization-theoretic approaches have been proposed to establish more principled rules for cloning and splitting: SteepGS \cite{wang2025steepest} leverages steepest-descent analysis to trigger Gaussian splits, whereas 3DGS-MCMC \cite{kheradmand20243d} formulates density control as a Markov Chain Monte Carlo (MCMC) sampling process with \(L_1\)-regularized pruning. More recently, Efficient Density Control \cite{deng2024efficient} introduces long-axis splitting operations and recovery-aware pruning strategies to reduce redundant Gaussians and accelerate convergence, while Gradient-Direction-Aware ADC \cite{zheng2025gradient} incorporates gradient coherence and angular alignment to more effectively determine which Gaussians should be cloned or split. However, as these methods are primarily tailored for static scene reconstruction, they tend to under-fit transiently visible regions and fail to jointly consider both geometric and textural discrepancies.

\begin{figure*}[ht]
	\centering
	\includegraphics[width=1.0\textwidth]{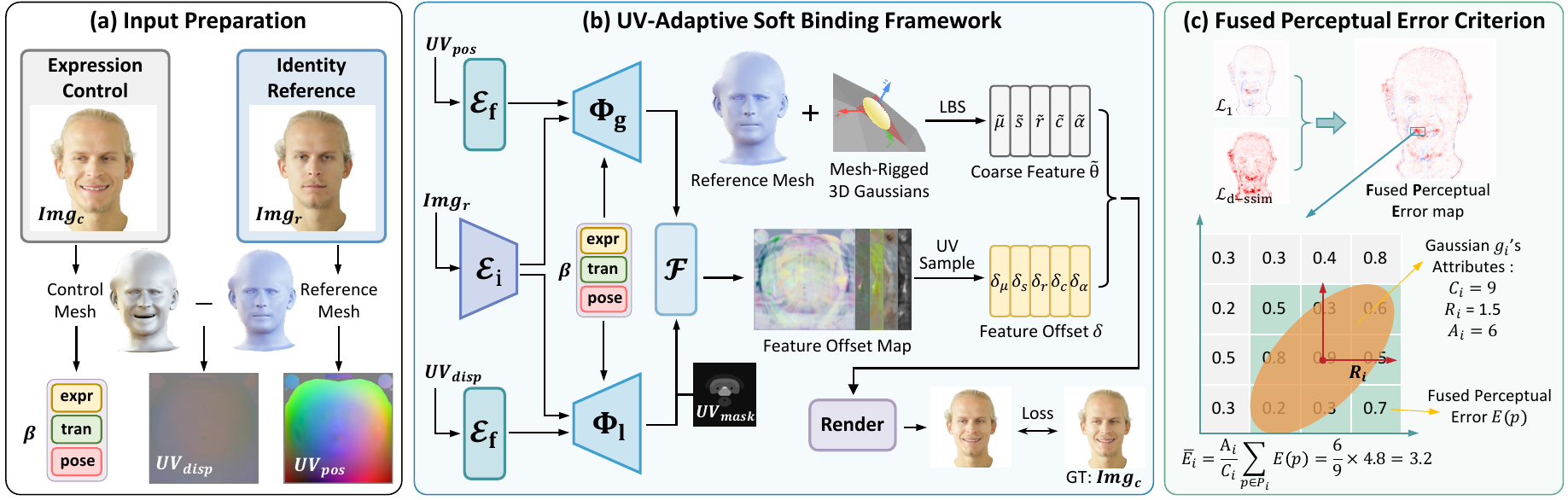}
	\caption{Overview of STAvatar. (a) In addition to a fixed identity reference image and its UV position map, we further rasterize the vertex offsets between reference mesh and control mesh to obtain a UV displacement map as input. (b) We construct a dual-branch network to predict a feature offset map in UV space, from which an offset \(\delta_i\) is sampled for each Gaussian \(g_i\). This offset is added to the coarsely estimated parameters \(\tilde{\theta}\) to get final parameters \(\theta^*\). The final images are then rendered using Gaussian Splatting. (c) We first construct a perceptual error map by combining \(\mathcal{L}_1\) map and \(\mathcal{L}_\mathrm{d\text{-}ssim}\) map. Then, we estimate the 2D projection of each Gaussian \(g_i\) using the recorded attributes, based on which the fused perceptual error is computed.}
	\vspace*{-8pt}
	\label{fig:method}
\end{figure*}
\section{Method}
We first introduce the general Gaussian avatar reconstruction pipeline in Sec.~\ref{preliminary}. Next, we present a UV-Adaptive Soft Binding framework in Sec.~\ref{soft}. The Temporal Adaptive Density Control strategy, which integrates FLAME‑Conditioned Temporal Clustering (FTC) and a Fused Perceptual Error with Average-Peak Criterion (FPE‑AP), is discussed in Sec.~\ref{temporal}. Finally, the detailed optimization process is described in Sec.~\ref{scheme}.

\subsection{Preliminaries}
\label{preliminary}
\textbf{Initialization. } Following \cite{qian2023gaussianavatars}, we initialize Gaussian primitives \(g_i\) based on the tracked FLAME model. Each \(g_i\) is bound to a mesh triangle and parameterized by a center position \(\bm{\mu}\), a scale matrix \(\bm{S}\), and a rotation matrix \(\bm{R}\) defined on the canonical FLAME mesh \cite{li2017learning}. The 3D Gaussian density is defined as:
\begin{equation}
    G(\bm{x}) = e^{-\frac{1}{2}(\bm{x} - \bm{\mu})^T \Sigma^{-1}(\bm{x} - \bm{\mu})},
\end{equation}
where \(\Sigma = \bm{R}\,\bm{S}\,\bm{S^\top}\,\bm{R^\top}\) to keep the covariance matrix positive. In addition to geometric parameters, each primitive also carries appearance attributes: opacity \(\alpha\) and color \(c\).

\noindent \textbf{Animate. } 
To animate these canonical primitives according to FLAME deformations, we first transform their parameters using the parent triangle’s barycentric mapping to get the coarsely estimated parameters \(\tilde\theta=\{\bm{\tilde\mu}, \bm{\tilde{s}}, \bm{\tilde{r}}, \tilde\alpha, \tilde{c}\}\). Specifically,
\begin{equation}
    \bm{\tilde{r}} = \bm{r} \bm{R},\ 
    \bm{\tilde{\mu}} = k \bm{r} \bm{\mu} + \bm{t},\ 
    \bm{\tilde{s}} = k \bm{s},\ 
    \tilde\alpha=\alpha, \ 
    \tilde{c}=c,
    \label{eq:lbs}
\end{equation}
where \(\bm{r}\) and \(\bm{t}\) denote the triangle’s relative rotation and barycenter translation, respectively, and the isotropic scale \(k\) reflects the triangle’s size.

\noindent \textbf{Render. } 
Given the final parameters \(\theta^{*}\) (see Sec.~\ref{soft}), the rendered color \(\mathbf{C}\) of a pixel is obtained by compositing all overlapping primitives in a depth-sorted compositing:
\begin{equation}
    \bm{C} = \sum_{i=1}^{N} c_i^*\,\alpha_i' \!\!\prod_{j=1}^{i-1}\!(1 - \alpha_j'),
    \label{eq:render}
\end{equation}
where \(c_i^*\) is the RGB color encoded via 3rd-degree spherical harmonics, and \(\alpha_i'\) is the 2D projected opacity obtained by evaluating the Gaussian’s contribution in image space. This compositing accounts for occlusion by sorting the primitives according to their depth values.

\subsection{UV-Adaptive Soft Binding Framework}
\label{soft}
Eq.~\eqref{eq:lbs} only yields a coarse deformation for each Gaussian \(g_i\), failing to capture fine details or appearance changes since opacity \(\tilde{\alpha}\) and color \(\tilde{c}\) remain constant under LBS. To address this while preserving the flexibility of ADC, we introduce the UV‑Adaptive Soft Binding framework.

\noindent \textbf{Input Preparation. }
As shown in Fig.~\ref{fig:method}(a), after tracking the flame model for video frames, we select a fixed reference image \(Img_r\) to extract the texture input and rasterize its UV position map \(UV_{pos}\) to obtain UV coordinate information. To further enhance the capacity for deformation modeling, we rasterize the vertex offsets between the reference mesh and the control mesh into UV space to generate an additional input \(UV_{disp}\).

\noindent \textbf{Soft Binding. } 
Fig.~\ref{fig:method}(b) illustrates our dual-branch network, which estimates per-Gaussian feature offsets in UV space to capture rich spatial context. We first extract texture features \(T\) by encoding \(Img_r\) with a U-Net encoder, denoted as \(\mathcal{E}_i\), such that \(T =\mathcal{E}_i(Img_r)\). These features are then concatenated with Fourier positional encoded position maps \(UV_{pos}^{'} = \mathcal{E}_f(UV_{pos})\) and displacement maps \(UV_{disp}^{'} =  \mathcal{E}_f(UV_{disp})\). Besides, both branches \(\Phi_g\) and \(\Phi_l\) take as input a control code \(\beta\), formed by concatenating expression, translation, and pose codes, to model Gaussian deformations. The decoded outputs, \(\omega_g\) and \(\omega_l\), are then fused to generate the final feature offset map \(\Delta_{map}\in \mathbb{R}^{256 \times 256 \times 13}\), where each texel stores a 13‑dimensional Gaussian offset. The entire  process can be described as:
\begin{gather}
    \omega_g = \Phi_g(T, UV_{pos}^{'}, \beta), \\
    \omega_l = \sum_{i=1}^{4} H_i(M_i \odot \Phi_l(T, UV_{disp}^{'}, \beta)), \\
    \Delta_{\!map} = \mathcal{F}(\omega_g, \omega_l),
\end{gather}
where \(M_i \in \{0,1\}^{256\times256}\) denotes the UV spatial mask for the \(i\)-th facial region, including the eyes, mouth, nose, and forehead, and \(\odot\) denotes the element-wise (Hadamard) product. The local branch \(\phi_l\) shares a common decoder across all regions and applies region-specific decoding heads \(H_i\) to the masked features. Note that \(Img_r\) can be selected randomly; in our experiments, we default to using the first frame.

\noindent \textbf{UV-Adaptive Sampling. } 
We sample each Gaussian \(g_i\)’s offset 
\(\delta_i = \{\delta_{\mu}, \delta_{s}, \delta_{r}, \delta_{\alpha}, \delta_{c}\}\) 
from the feature offset map \(\Delta_{map}\) by assigning each \(g_i\) a UV coordinate via our UV-Adaptive Sampling, which is updated upon densification to accommodate dynamic changes in Gaussian count. A brief algorithmic description of this sampling procedure is provided in Algorithm~\ref{alg:sample}, with a detailed version included in the supplementary material. After sampling, we apply different activation functions to ensure numerical stability and constrain the effect of the predicted offsets \(\delta_i\) (details in Appendix). The final Gaussian parameters \(\theta^*\) are then computed as:
\begin{equation}
\begin{aligned}
    \mu^* &= \tilde{\mu} + \delta_{\mu}, \quad
    c^* = \tilde{c} + \delta_{c}, \quad
    \alpha^* = \tilde{\alpha} + \delta_{\alpha}, \\
    s^* &= \tilde{s} \odot \delta_{s}, \quad
    r^* = q(\tilde{r}, \delta_{r}),
\end{aligned}
\end{equation}
where \(q(\cdot, \cdot)\) represents the Hamilton product between two unit quaternions, used to compose the rotation \(\tilde{r}\) and its predicted offset \(\delta_{r}\). After obtaining the final parameters \(\theta^*\), we employ Eq.~\eqref{eq:render} to render the images.

\begin{algorithm}[!t]
\caption{UV-Adaptive Sampling}
\label{alg:sample}
\textbf{Input}: Face bindings $\mathcal{B}$, UV vertices $\mathcal{V}_{uv}$, UV faces $\mathcal{F}_{uv}$\\
\textbf{Output}: $\mathcal{U} \in \mathbb{R}^{N \times 2}$: UV coordinates per Gaussian\\
\begin{algorithmic}[1]
\STATE $(\mathcal{M}_{face}, \mathcal{M}_{zbary}) \gets \text{Rasterize}(\mathcal{V}_{uv}, \mathcal{F}_{uv})$
\STATE Build per-face pixel pool $\mathcal{P}$ from $\mathcal{M}_{face}$
\STATE $\mathcal{F}_{valid} \gets \{\,f \in \mathcal{F}_{uv} \mid \mathcal{P}[f] \neq \emptyset \,\}$
\STATE Initialize empty barycentric buffer $\mathbf{b}$ and face index buffer $\mathbf{f}_{idx}$
\FOR{each face $f$ in face bindings $\mathcal{B}$}
    \STATE $C_f \gets$ count of points bound to face $f$
    \STATE $\mathcal{I}_f \gets$ indices of points bound to face $f$
    \IF{$f \in \mathcal{F}_{valid}$} 
        \STATE $\mathbf{p}_f \gets$ pixel list for face $f$ from $\mathcal{P}$
        \STATE Sample $C_f$ pixels from $\mathbf{p}_f$
        \STATE Extract corresponding barycentric coordinates $\mathbf{b}_f$
    \ELSE 
        \STATE Analytically sample $C_f$ barycentric coordinates $\mathbf{b}_f$
    \ENDIF
    \STATE Assign $\mathbf{b}_f$ to $\mathbf{b}[\mathcal{I}_f]$, and fill $\mathbf{f}_{idx}[\mathcal{I}_f] \gets f$
\ENDFOR
\STATE $\mathcal{U} \gets \text{BarycentricReweight}(\mathcal{V}_{uv}, \mathcal{F}_{uv}, \mathbf{f}_{idx}, \mathbf{b})$
\end{algorithmic}
\end{algorithm}

\subsection{Temporal Adaptive Density Control}
\label{temporal}
\textbf{FPE‑AP. } 
To identify Gaussians that require cloning, we introduce a more interpretable and comprehensive criterion, FPE-AP, which directly indicates regions exhibiting significant geometric or texture errors that may benefit from more Gaussian primitives. Since the error at each pixel is influenced by multiple overlapping Gaussians, it is computationally intractable and inefficient to compute the exact contribution of each individual Gaussian.

To address this, as illustrated in Fig.~\ref{fig:method}(c), we first construct a fused perceptual error map:
\begin{equation}
    E = (1-\lambda_1)\,\left\lvert\mathcal{L}_1\right\rvert + \lambda_1\,\mathcal{L}_{\mathrm{d\text{-}ssim}},
\end{equation}
where \(\mathcal{L}_1\) denotes the per-pixel absolute difference, and \(\mathcal{L}_{\mathrm{d\text{-}ssim}}\) represents the per-pixel dissimilarity, derived from the Structural Similarity Index (SSIM). We set \(\lambda_1=0.2\) to keep it consistent with the RGB loss design in Eq.~\eqref{eq:rgb}, which better aligns with human perceptual sensitivity.

To estimate each Gaussian’s 2D projection for fused perceptual error calculation, we record its screen-space center \((x_i, y_i)\), total number of covered pixels \(C_i\), and accumulated alpha blending weight \(A_i = \sum_{p \in \mathcal{P}_i} a_i(p)\), where \(a_i(p)\) is the blending weight of \(g_i\) at pixel \(p\). The influence region is defined as a square centered at \((x_i, y_i)\) with half-extent \(R_i = \left\lfloor \frac{\sqrt{C_i}}{2} \right\rfloor\), and the corresponding pixel set \(\mathcal{P}_i\) includes all pixels within this region. We then approximate the average fused perceptual error for \(g_i\) as
\begin{equation}
    \bar{E}_i = \frac{A_i}{C_i} \sum_{p \in \mathcal{P}_i} E(p),
\end{equation}
where all window-sum computations are efficiently accelerated using a two-dimensional summed-area table (details in the supplementary material).

Furthermore, to capture transient spikes, we define a peak fused perceptual error across all iterations \(t\) as
\begin{equation}
E_i^{\mathrm{peak}} = \max_t \left( \frac{A_i^{(t)}}{C_i^{(t)}} \sum_{p \in \mathcal{P}_i^{(t)}} E^{(t)}(p) \right),
\end{equation}
and select top 3\% Gaussians by \(E_i^{\mathrm{peak}}\) to form the set \(\mathcal{S}_{\mathrm{peak}}\). Finally, Gaussian \(g_i\) is cloned if
\begin{equation}
\bar{E}_i > \tau_{\mathrm{avg}} \quad \text{or} \quad i \in \mathcal{S}_{\mathrm{peak}},
\end{equation}
where \(\tau_{\mathrm{avg}} = 1 \times 10^{-3}\). Note that, for Gaussian splitting, we still use the positional gradient as criterion since splitting is primarily driven by geometric inconsistency.

\begin{figure}[!t]
    \centering
    \includegraphics[width=1\linewidth]{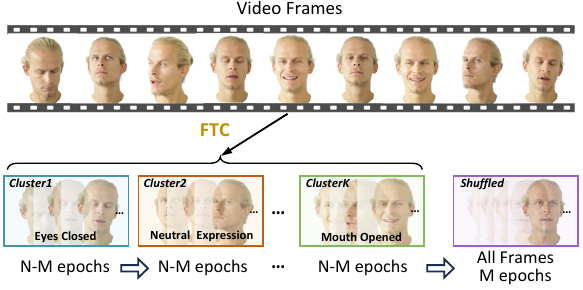}
    \caption{FLAME‑Conditioned Temporal Clustering. We cluster video frames into \(K\) clusters and conduct ADC within each cluster's training.}
    \vspace*{-5pt}
    \label{fig:ftc}
\end{figure}

\noindent \textbf{FTC. }
As shown in Fig.~\ref{fig:ftc}, we cluster video frames into \(K\) clusters based on their FLAME parameters (expression, pose, and translation), with respective weights of 0.3, 0.6, and 0.1. Before clustering, we apply Principal Component Analysis (PCA) to reduce the parameter dimensionality for more effective distance computation. We then perform K-means clustering, which ensures the densification criterion is computed among structurally similar frames. This strategy mitigates insufficient densification in transiently visible regions. To accommodate different identities, we select \(K\) within \([5,12]\) by maximizing the average silhouette score.

During training, each cluster is first optimized for \(N{-}M\) epochs, during which ADC is applied. After all clusters have been trained, we conduct additional \(M\) epochs using randomly shuffled data to mitigate potential inconsistencies introduced by inter-cluster variations. In our experiments, we set \(N=6\) and \(M=1\).

\begin{table*}[htb]
	\centering
	\caption{Comparison of quantitative results with state-of-the-art methods. The \colorbox{c1}{best} and \colorbox{c2}{second} are highlighted, respectively.}
	\vspace*{-5pt}
	\resizebox{\textwidth}{!}
	{
		\fontsize{9pt}{11pt}\selectfont
		\begin{tabular}{ccccccccccccc}
			\toprule
			\multirow{2}{*}{Method} & \multicolumn{3}{c}{INSTA Dataset} & \multicolumn{3}{c}{PointAvatar Dataset} & \multicolumn{3}{c}{NerFace Dataset} & \multicolumn{3}{c}{HDTF Dataset} \\
			\cmidrule(lr){2-13}
			& PSNR↑ & SSIM↑ & LPIPS↓ & PSNR↑ & SSIM↑ & LPIPS↓ & PSNR↑ & SSIM↑ & LPIPS↓ & PSNR↑ & SSIM↑ & LPIPS↓ \\
			\midrule
			SA \cite{SplattingAvatar:CVPR2024} & 27.48 & 0.9329 & 0.1046 & 24.93 & 0.8938 & 0.1478 & 26.14 & 0.9238 & 0.1145 & 26.02 & 0.8922 & 0.1821 \\
			MGA \cite{chen2024monogaussianavatar} & 27.34 & 0.9351 & 0.0859 & 27.91 & 0.9277 & 0.1113 & 26.52 & 0.9333 & 0.0900 & 27.04 & 0.9171 & 0.1225 \\
			GA \cite{qian2023gaussianavatars} & 26.98 & 0.9378 & 0.0851 & 24.62 & 0.9073 & 0.1333 & 25.74 & 0.9279 & 0.0965 & 25.08 & 0.8879 & 0.1763 \\
			FA \cite{xiang2024flashavatar} & 27.90 & 0.9357 & 0.0563 & 26.19 & 0.9055 & 0.0947 & 26.96 & 0.9331 & 0.0555 & 26.83 & 0.9110 & \cellcolor{c2}0.0774 \\
			RGBA \cite{li2025rgbavatar} & \cellcolor{c2}28.41 & \cellcolor{c2}0.9493 & 0.0518 & 26.67 & 0.9217 & 0.0883 & \cellcolor{c2}27.13 & \cellcolor{c2}0.9438 & \cellcolor{c2}0.0550 & 26.72 & \cellcolor{c2}0.9277 & 0.0821 \\
			Fate \cite{zhang2025fate} & 28.33 & 0.9446 & \cellcolor{c2}0.0508 & \cellcolor{c1}28.36 & \cellcolor{c2}0.9287 & \cellcolor{c2}0.0776 & 27.12 & 0.9364 & 0.0558 & \cellcolor{c2}27.18 & 0.9187 & 0.0849 \\
			\midrule
			Ours & \cellcolor{c1}30.63 & \cellcolor{c1}0.9587 & \cellcolor{c1}0.0304
			& \cellcolor{c2}28.25 & \cellcolor{c1}0.9337 & \cellcolor{c1}0.0495 & \cellcolor{c1}30.08 & \cellcolor{c1}0.9567 & \cellcolor{c1}0.0311 & \cellcolor{c1}27.99 & \cellcolor{c1}0.9315 & \cellcolor{c1}0.0542 \\
			\bottomrule
		\end{tabular}
	}
	\label{tab:recon}
\end{table*}

\begin{figure*}[!ht]
	\centering
	\includegraphics[width=1.0\textwidth]{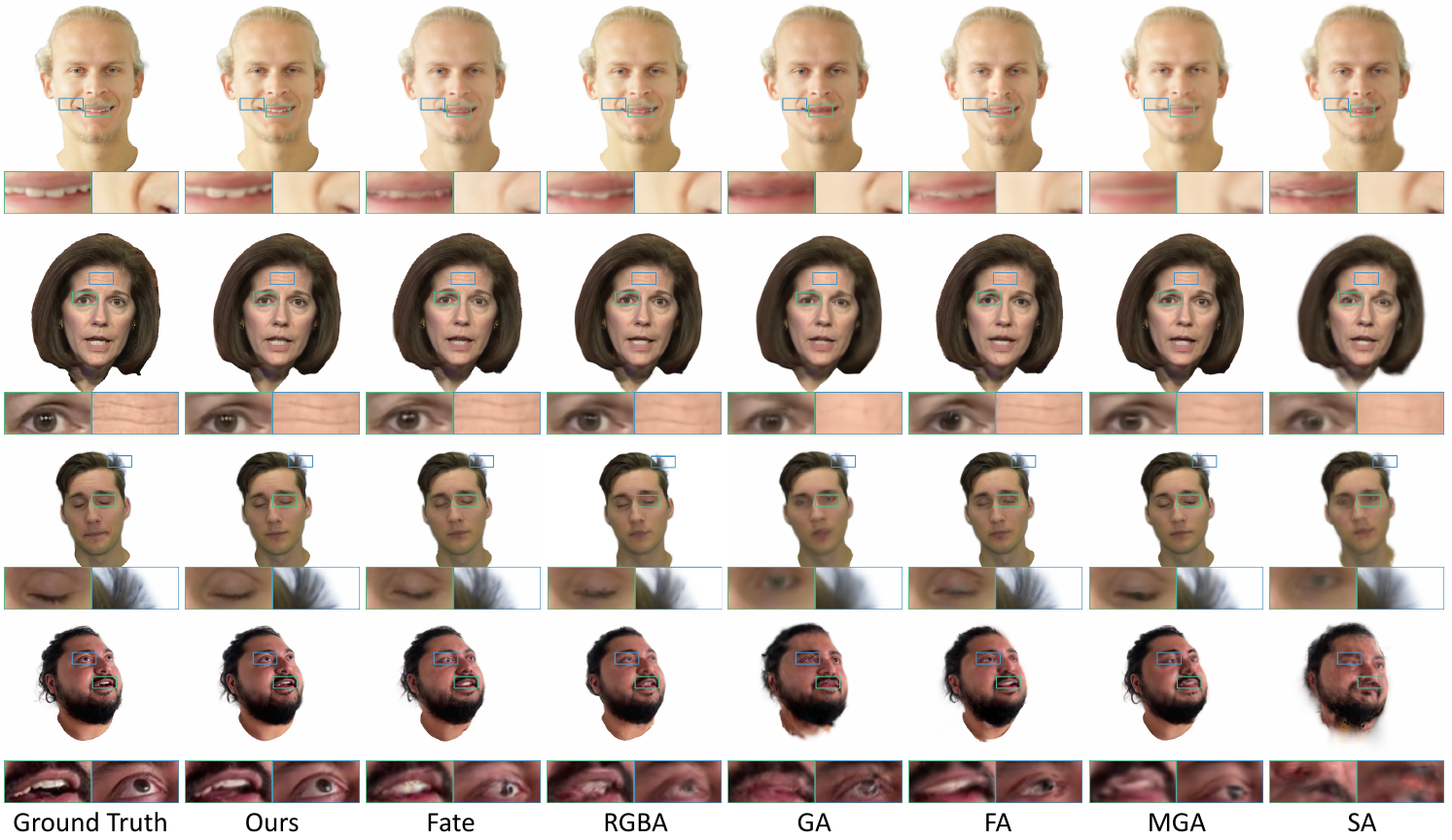}
	\caption{Qualitative results of head avatar reconstruction. Our method recovers finer details and delicate structures such as wrinkles and teeth. Moreover, it produces clearer results in challenging regions like mouth interiors and eyelids.}
	\vspace*{-5pt}
	\label{fig:recon}
\end{figure*}

\subsection{Training Objective and Optimization}
\label{scheme}
\noindent \textbf{RGB Loss. } 
We supervise the rendered images using a combination of \(\mathcal{L}_1\) term and D-SSIM term following \cite{kerbl3Dgaussians}. To better capture high-frequency facial details, we further introduce the perceptual loss \(\mathcal{L}_\mathrm{vgg}\), which is activated only in the second half of training to balance performance and efficiency. The overall image loss is defined as:
\begin{equation}
    \mathcal{L}_\mathrm{rgb} = (1 - \lambda_1) \mathcal{L}_1 + \lambda_1 \mathcal{L}_{\mathrm{d\text{-}ssim}} + \gamma \lambda_2 \mathcal{L}_\mathrm{vgg},
    \label{eq:rgb}
\end{equation}
where \(\lambda_1 = 0.2\), \(\lambda_2 = 0.05\), and \(\gamma = 1\) only during the latter half of training.

\noindent \textbf{Regularization Loss. } 
To regularize the predicted offset \(\delta_i\), we apply a regularization term to constrain the scale and color offsets:
\begin{equation}
    \mathcal{L}_{\mathrm{offset}} = \lambda_3 |\delta_s - 1| + \lambda_4 \delta_c,
\end{equation}
where \(\lambda_3=0.01\) and \(\lambda_4=0.001\). Following GaussianAvatars \cite{qian2023gaussianavatars}, we also employ a position loss and a scale loss to regularize the spatial alignment and local scale of the 3D Gaussians with respect to their parent triangles. 

The total training loss is thus defined as:
\begin{equation}
    \mathcal{L} = \mathcal{L}_\mathrm{rgb} + \mathcal{L}_\mathrm{offset} + \lambda_p \mathcal{L}_\mathrm{position} + \lambda_s \mathcal{L}_\mathrm{scale},
\end{equation}
where \(\lambda_p = 0.01\) and \(\lambda_s = 1\). 

\noindent \textbf{Optimization. } 
We employ Adam \cite{kingma2014adam} for parameter optimization, using identical hyperparameter settings across all subjects. The learning rate is set to \(1\times10^{-4}\) for the UV-Adaptive Soft Binding framework, while the learning rates for the remaining parameters follow those used in 3D Gaussian Splatting. In addition to the Gaussian Splatting parameters, we also finetune the expression, translation and joint rotation parameters of FLAME for each timestep as done in GaussianAvatars. As each Gaussian is bound to a mesh triangle, we do not observe noticeable floating Gaussians and thus skip opacity resetting in ADC.
\section{Experiments}

\subsection{Experimental Setup}
\label{sec:settings}
\begin{figure}[t]
	\centering
	\includegraphics[width=1.0\linewidth]{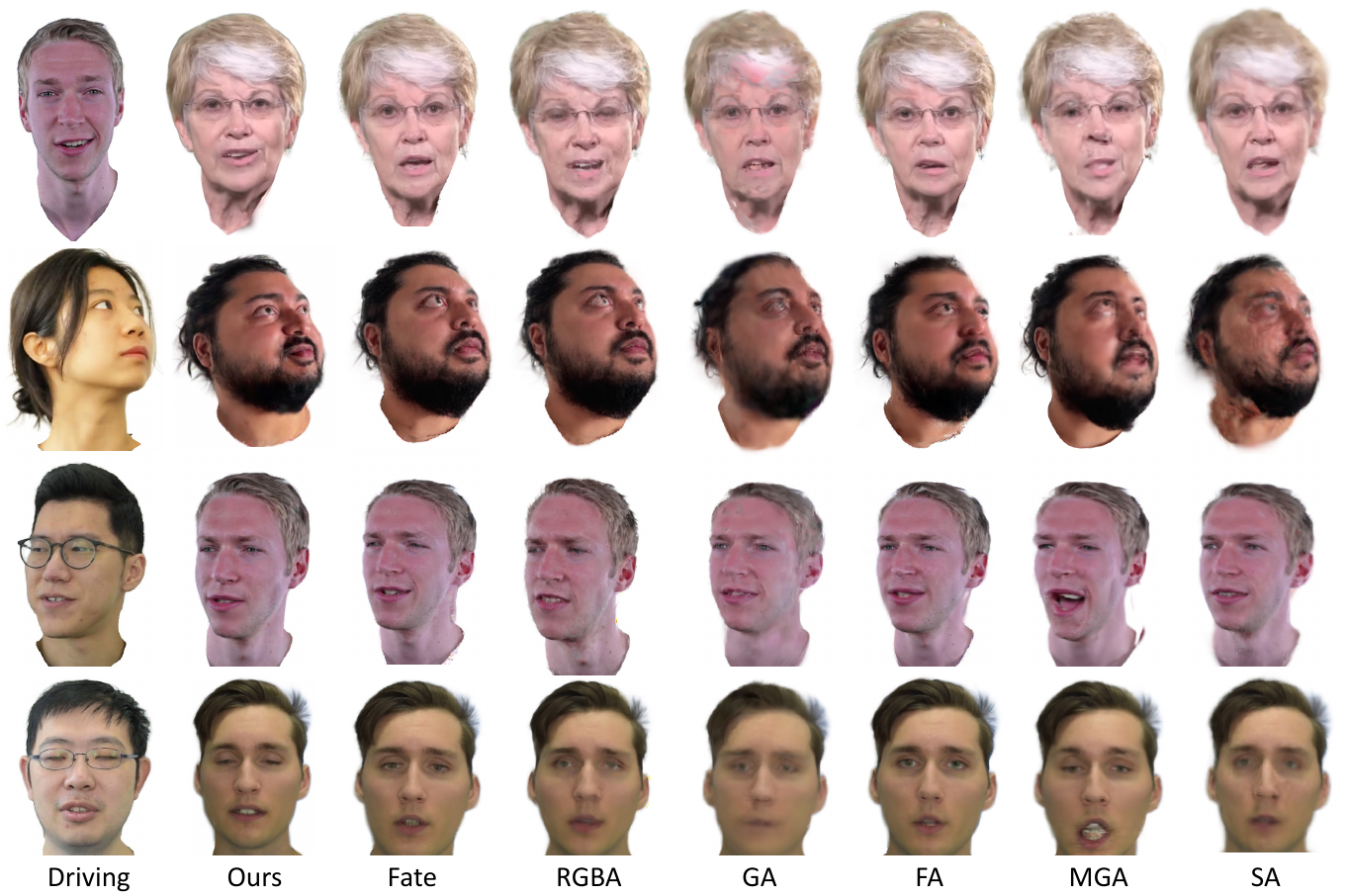}
	\caption{Qualitative results of cross-identity reenactment. Our method accurately animates the source avatar performing expressions such as smiling and eye-closing.}
	\vspace*{-5pt}
	\label{fig:cross}
\end{figure}

\textbf{Implementation Details. } 
We conduct both training and inference on a single NVIDIA RTX 3090 GPU. Each identity's video is trained for 6 epochs. All videos are cropped and resized to a resolution of \(512\times512\). A robust matting method \cite{lin2022robust} is applied to extract the foreground with the background set to white. For avatar initialization, we adopt VHAP \cite{qian2024vhap} for monocular head tracking.

\noindent \textbf{Dataset. } 
We evaluate our method on four datasets, including INSTA \cite{INSTA:CVPR2023}, PointAvatar \cite{Zheng2023pointavatar}, Nerface \cite{gafni2021dynamic}, and HDTF \cite{zhang2021flow}, comprising a total of 22 identities. For PointAvatar, we follow the original data split strategy in \cite{Zheng2023pointavatar}, while for the other datasets, the last 450 frames are reserved as the test set.

\noindent \textbf{Baselines. }
We compare our method against six state-of-the-art Gaussian avatar reconstruction approaches: GaussianAvatars (GA) \cite{qian2023gaussianavatars}, FateAvatar (Fate) \cite{zhang2025fate}, RGAvatar (RGBA) \cite{li2025rgbavatar}, FlashAvatar (FA) \cite{xiang2024flashavatar}, MonoGaussianAvatar (MGA) \cite{chen2024monogaussianavatar}, and SplattingAvatar (SA) \cite{SplattingAvatar:CVPR2024}. Note that simply equalizing or increasing training epochs across methods may not yield optimal results. Accordingly, we follow the settings in the original papers, setting 10 epochs for GA, Fate, and RGBA, 20 for FA, 30 for SA, and 100 for MGA to allow each method to converge properly. Since GA is originally designed for multi-view reconstruction, we adopt various settings from original papers for fair comparison.

\subsection{Evaluation}
\textbf{Quantitative Results. } 
We select PSNR, SSIM, and LPIPS \cite{zhang2018unreasonable} as evaluation metrics to comprehensively assess the visual quality of the reconstructed avatars. As shown in Tab.~\ref{tab:recon}, our method consistently outperforms existing state-of-the-art approaches across most datasets and metrics, achieving notable improvements in reconstruction quality. In particular, our method attains the highest SSIM scores and the lowest LPIPS values on all four datasets, demonstrating its strong ability to preserve both geometric accuracy and perceptual fidelity.

\noindent \textbf{Qualitative Results. } 
As shown in Fig.~\ref{fig:recon}, owing to the non-rigid and flexible deformation capability of our UV-Adaptive Soft Binding framework, our method effectively recovers fine-grained details and subtle structures, such as wrinkles and hair. Furthermore, benefiting from the Temporal ADC strategy, our method can produce clearer and higher-quality reconstructions in challenging regions like mouth interiors and eyelids, which are only visible in a subset of frames and are typically difficult to reconstruct. We further evaluate our method on cross-identity reenactment as a real-world test for avatar animation.  As illustrated in Fig.~\ref{fig:cross}, our method accurately animates the source avatar performing expressions such as smiling and eye-closing, while faithfully preserving the identity-specific characteristics and fine-grained details.

\begin{table}[t]
	\centering
	\caption{Ablation quantitative results on the INSTA dataset.}
	\vspace*{-5pt}
	{
		\begin{tabular}{llll}
			\toprule
			Method             & PSNR↑ & SSIM↑  & LPIPS↓ \\ \midrule
			w/o Soft Binding   & 29.66 & 0.9525 & 0.0398 \\
			w/o ADC            & 30.31 & 0.9529 & 0.0505 \\
			w/o FPE-AP         & 30.38 & 0.9574 & 0.0321 \\
			w/o FTC            & 30.43 & 0.9579 & 0.0310 \\
			w/o Local Decoder \(\phi_l\)  & 30.36 & 0.9575 & \cellcolor{c2}0.0306 \\
			w/o \(\mathcal{L}_{\mathrm{vgg}}\)    & \cellcolor{c2}30.58 & \cellcolor{c1}0.9592 & 0.0426 \\
			\midrule
			Ours               & \cellcolor{c1}30.63 & \cellcolor{c2}0.9587 & \cellcolor{c1}0.0304 \\
			\bottomrule
		\end{tabular}
	}
	\label{tab:ablation}
\end{table}

\begin{figure}[t]
	\centering
	\includegraphics[width=1\linewidth]{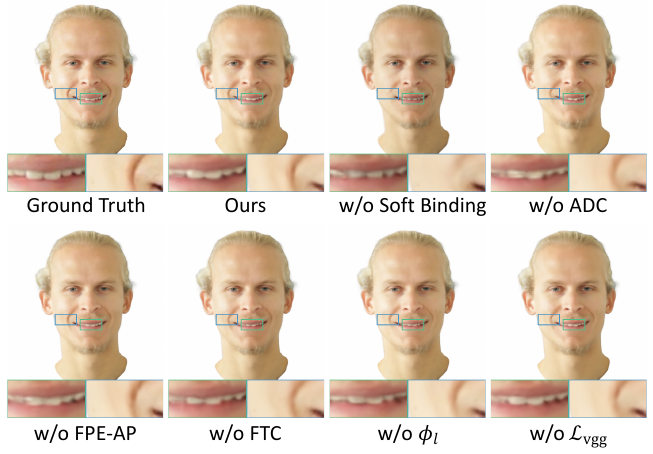}
	\caption{Qualitative results of the ablation study on marcel case. Our full method produces more realistic textures and geometric structures in transiently visible regions, such as the mouth interior, and captures more fine details.}
	\vspace{-5pt}
	\label{fig:ablation}
\end{figure}

\subsection{Ablation Study}
We conduct an ablation study on self-reenactment using the INSTA dataset. The quantitative results are presented in Tab.~\ref{tab:ablation}, while the qualitative comparisons are shown in Fig.~\ref{fig:ablation}, demonstrating the effectiveness of each component. Additionally, we provide the ablation analysis of all hyper-parameters in the supplementary materials.

\begin{figure}[t]
	\centering
	\includegraphics[width=1\linewidth]{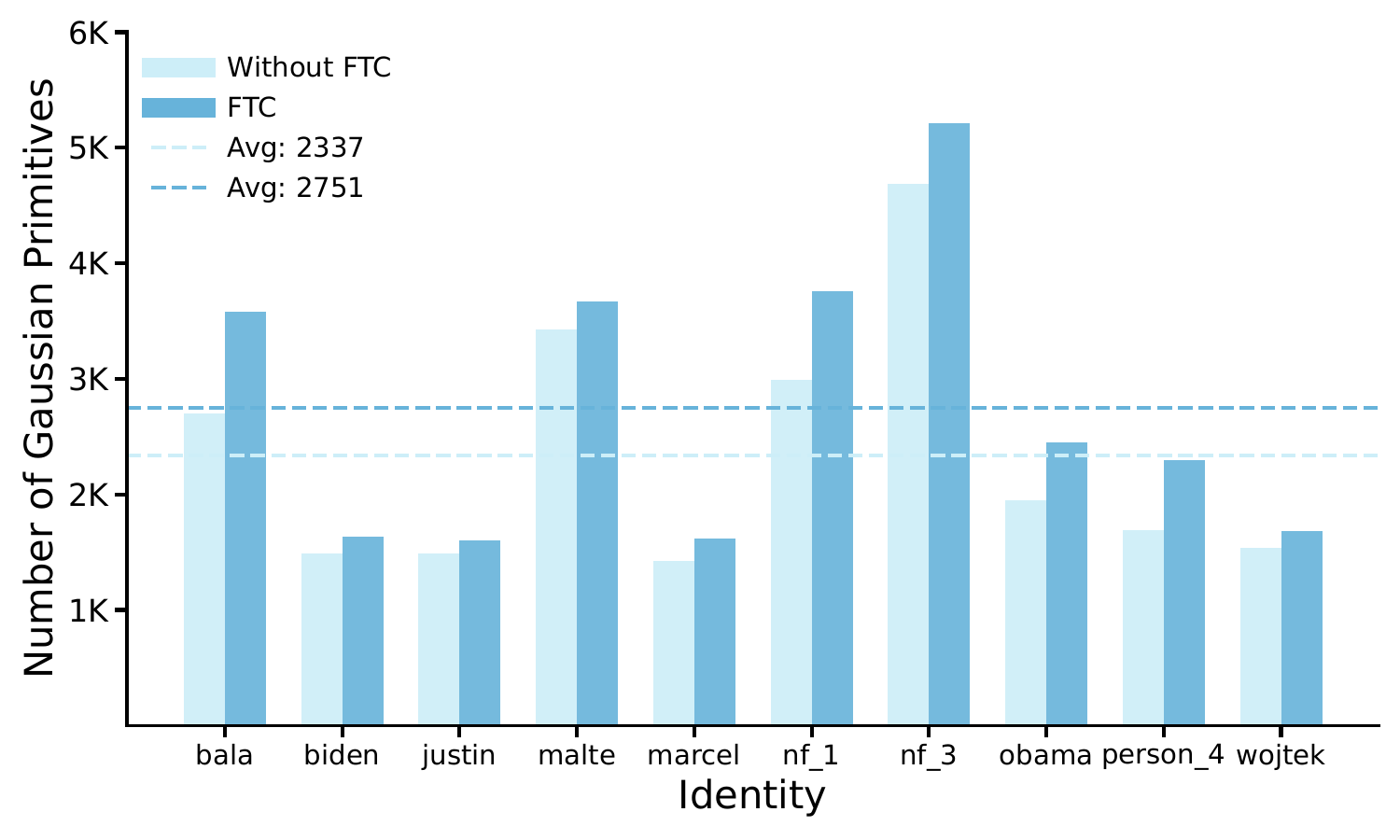}
	\caption{Comparison of Gaussian primitive counts within the mouth interior between the vanilla training strategies and FTC.}
	\vspace{-5pt}
	\label{fig:ftc_point}
\end{figure}

\noindent \textbf{Without Soft Binding. } 
As shown in Tab.~\ref{tab:ablation} and Fig.~\ref{fig:ablation}, removing the Soft Binding framework results in a noticeable degradation across all metrics and visibly poorer reconstruction quality. Relying solely on LBS-based deformation substantially weakens the modeling capability, thereby failing to capture fine-grained details and resulting in blurred teeth as well as the absence of smile-induced facial wrinkles.

\noindent \textbf{Without ADC. } 
Removing ADC leads to a significant increase in the LPIPS score, suggesting a degradation in perceptual quality due to the loss of high-frequency details. As shown in Fig.~\ref{fig:ablation}, the teeth region and facial wrinkles are insufficiently reconstructed by Gaussians, resulting in overly smooth and less detailed renderings.

\noindent \textbf{Without FPE-AP. } 
Replacing FPE-AP criterion with vanilla positional gradient leads to blurred teeth renderings, as the latter considers only geometric information and fails to add Gaussians in regions with high-frequency details.

\noindent \textbf{Without FTC. } 
For the vanilla training strategy, we set the densification interval based on the number of clusters to ensure that the total number of densification steps matches that of FTC. The teeth region exhibits incomplete reconstruction due to its transiently visible nature.

\noindent \textbf{Without Local Decoder \(\phi_l\). }
In this setting, only the global branch is used to model the Gaussian deformation field. As illustrated in Fig.~\ref{fig:ablation}, facial wrinkles are reconstructed less accurately, and the teeth region appears blurrier.

\noindent \textbf{Without \(\mathcal{L}_{\mathrm{vgg}}\). }
As shown in Tab.~\ref{tab:ablation}, removing the perceptual loss \(\mathcal{L}_\mathrm{vgg}\) improves SSIM marginally but significantly increases LPIPS score, reflecting overly smooth textures.

\begin{figure}[h]
	\centering
	\includegraphics[width=0.98\linewidth]{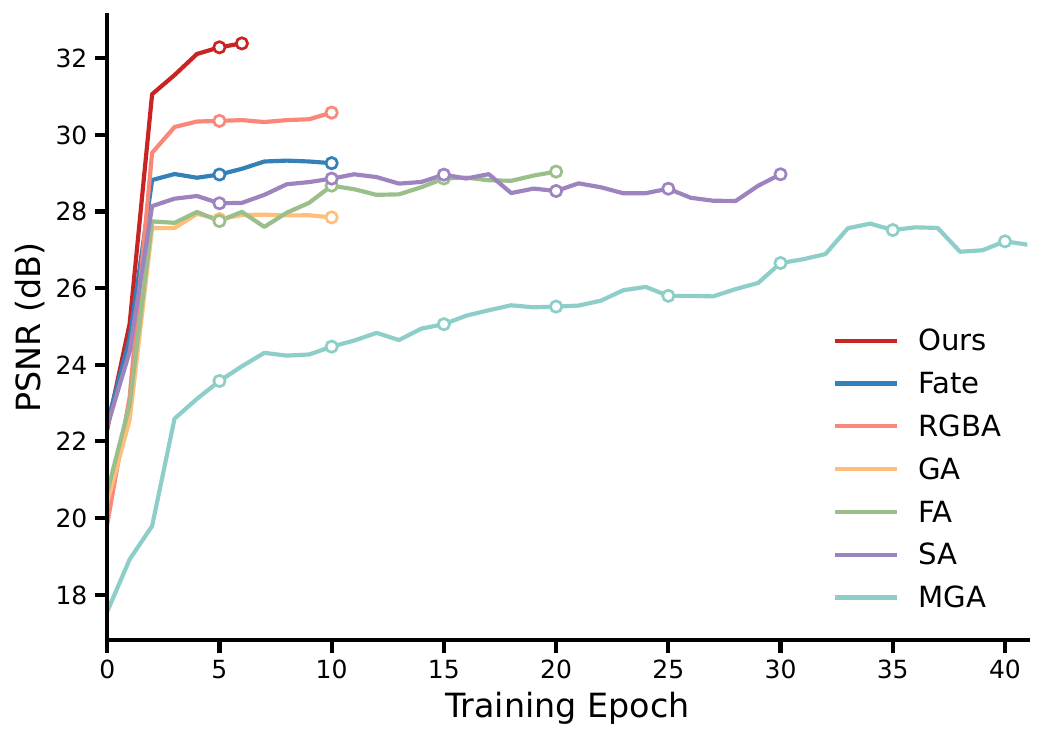}
	\caption{Comparison of PSNR performance across different methods during training. STAvatar exhibits the highest training efficiency among all compared approaches.}
	\vspace{-5pt}
	\label{fig:eff}
\end{figure}

\subsection{Gaussian Number Comparison} 
To further validate the effectiveness of FTC in transiently visible or frequently occluded regions, we count the number of Gaussian primitives whose UV coordinates fall within the mouth region mask. As illustrated in Fig.~\ref{fig:ftc_point}, we compare the number of Gaussian primitives in the mouth interior with and without FTC on the INSTA dataset. Although some variations exist across identities due to differences in motion and expression within video frames, it is evident that all identities exhibit increased density in the mouth region when using FTC, with an average gain of over 400 primitives (approximately 17\%).

\subsection{Training Efficiency Analysis}
As shown in Fig.~\ref{fig:eff}, we present the training progress and corresponding performance measured in terms of PSNR for different methods. The training epochs follow the settings described in Sec.~\ref{sec:settings} to ensure that all methods fully converge. For MonoGaussianAvatar, only the first 40 epochs are reported because its PSNR shows negligible improvement in the later stages of training. In addition, since different methods adopt distinct initialization strategies, their performance at epoch 0 varies accordingly. Notably, our method, STAvatar, exhibits the highest training efficiency among all compared approaches and almost reaches convergence within 6 epochs, surpassing the second-best method by over 2 dB.
\section{Conclusion}
In this paper, we present STAvatar, a novel method for high-fidelity and training-efficient reconstruction of animated avatars from monocular videos. The proposed UV-Adaptive Soft Binding framework enables flexible and non-rigid deformation field modeling while remaining compatible with the Adaptive Density Control (ADC), thereby effectively capturing subtle expressions and fine-grained details. Besides, our Temporal ADC strategy addresses the limitations of vanilla ADC in dynamic avatar reconstruction, resulting in more accurate and complete reconstructions in frequently occluded regions. Extensive experiments demonstrate that our method significantly outperforms previous approaches in both reconstruction quality and training efficiency.
\section*{Acknowledgement}
This work was supported in part by Beijing Natural Science Foundation L242092, Chinese National Natural Science Foundation Projects 92570119, 62276254, U23B2054, the Science and Technology Development Fund of Macau Project 0140/2024/AGJ, and InnoHK program.
{
    \small
    \bibliographystyle{ieeenat_fullname}
    \bibliography{main}

@String(CVPR= {IEEE Conf. Comput. Vis. Pattern Recog.})

@String(TOG= {ACM Trans. Graph.})

@String(CVPR  = {CVPR})

@String(TOG   = {ACM TOG})

@article{kerbl3Dgaussians,
      author       = {Kerbl, Bernhard and Kopanas, Georgios and Leimk{\"u}hler, Thomas and Drettakis, George},
      title        = {3D Gaussian Splatting for Real-Time Radiance Field Rendering},
      journal      = {ACM Transactions on Graphics},
      number       = {4},
      volume       = {42},
      month        = {July},
      year         = {2023},
      url          = {https://repo-sam.inria.fr/fungraph/3d-gaussian-splatting/}
}

@article{qian2023gaussianavatars,
    title={GaussianAvatars: Photorealistic Head Avatars with Rigged 3D Gaussians},
    author={Qian, Shenhan and Kirschstein, Tobias and Schoneveld, Liam and Davoli, Davide and Giebenhain, Simon and Nie\ss{}ner, Matthias},
    journal=CVPR,
    year={2024}
}

@inproceedings{xiang2024flashavatar,
    author    = {Jun Xiang and Xuan Gao and Yudong Guo and Juyong Zhang},
    title     = {FlashAvatar: High-fidelity Head Avatar with Efficient Gaussian Embedding},
    booktitle = CVPR,
    year      = {2024},
}

@inproceedings{SplattingAvatar:CVPR2024,
    title = {{SplattingAvatar: Realistic Real-Time Human Avatars with Mesh-Embedded Gaussian Splatting}},
    author = {Shao, Zhijing and Wang, Zhaolong and Li, Zhuang and Wang, Duotun and Lin, Xiangru and Zhang, Yu and Fan, Mingming and Wang, Zeyu},
    booktitle = CVPR,
    year = {2024}
}

@inproceedings{chen2024monogaussianavatar,
    title={Monogaussianavatar: Monocular gaussian point-based head avatar},
    author={Chen, Yufan and Wang, Lizhen and Li, Qijing and Xiao, Hongjiang and Zhang, Shengping and Yao, Hongxun and Liu, Yebin},
    booktitle={ACM SIGGRAPH 2024 Conference Papers},
    year={2024}
}

@inproceedings{zhang2025fate,
	title={FATE: Full-head Gaussian Avatar with Textural Editing from Monocular Video}, 
	author={Zhang, Jiawei and Wu, Zijian and Liang, Zhiyang and Gong, Yicheng and Hu, Dongfang and Yao, Yao and Cao, Xun and Zhu, Hao},
	booktitle={Proceedings of the IEEE/CVF Conference on Computer Vision and Pattern Recognition},
	year={2025},
}

@misc{qian2024vhap,
  title={VHAP: Versatile Head Alignment with Adaptive Appearance Priors},
  author={Qian, Shenhan},
  year={2024},
  month={sep},
  doi={10.5281/zenodo.14988309},
  url={https://github.com/ShenhanQian/VHAP}
}

@inproceedings{INSTA:CVPR2023,
  title = {Instant Volumetric Head Avatars},
  author = {Zielonka, Wojciech and Bolkart, Timo and Thies, Justus},
  booktitle = {Conference on Computer Vision and Pattern Recognition},
  year = {2023}
}

@inproceedings{Zheng2023pointavatar,
  author    = {Yufeng Zheng and Wang Yifan and Gordon Wetzstein and Michael J. Black and Otmar Hilliges},
  title     = {PointAvatar: Deformable Point-based Head Avatars from Videos},
  booktitle = {Proceedings of the IEEE/CVF Conference on Computer Vision and Pattern Recognition (CVPR)}, 
  year = {2023}
}

@inproceedings{gafni2021dynamic,
  title={Dynamic neural radiance fields for monocular 4d facial avatar reconstruction},
  author={Gafni, Guy and Thies, Justus and Zollhofer, Michael and Nie{\ss}ner, Matthias},
  booktitle={Proceedings of the IEEE/CVF Conference on Computer Vision and Pattern Recognition},
  pages={8649--8658},
  year={2021}
}

@inproceedings{HRAvatar,
	author    = {Zhang, Dongbin and Liu, Yunfei and Lin, Lijian and Zhu, Ye and Chen, Kangjie and Qin, Minghan and Li, Yu and Wang, Haoqian},
	title     = {HRAvatar: High-Quality and Relightable Gaussian Head Avatar},
	booktitle = {Proceedings of the Computer Vision and Pattern Recognition Conference (CVPR)},
	month     = {June},
	year      = {2025},
	pages     = {26285-26296}
}

@article{GUAVA,
	title={GUAVA: Generalizable Upper Body 3D Gaussian Avatar},
	author={Zhang, Dongbin and Liu, Yunfei and Lin, Lijian and Zhu, Ye and Li, Yang and Qin, Minghan and Li, Yu and Wang, Haoqian},
	journal={arXiv preprint arXiv:2505.03351},
	year={2025}
}

@inproceedings{zhang2021flow,
  title={Flow-guided one-shot talking face generation with a high-resolution audio-visual dataset},
  author={Zhang, Zhimeng and Li, Lincheng and Ding, Yu and Fan, Changjie},
  booktitle={Proceedings of the IEEE/CVF conference on computer vision and pattern recognition},
  pages={3661--3670},
  year={2021}
}

@article{li2017learning,
  title={Learning a model of facial shape and expression from 4D scans.},
  author={Li, Tianye and Bolkart, Timo and Black, Michael J and Li, Hao and Romero, Javier},
  journal={ACM Trans. Graph.},
  volume={36},
  number={6},
  pages={194--1},
  year={2017}
}

@article{feng2021learning,
  title={Learning an animatable detailed 3D face model from in-the-wild images},
  author={Feng, Yao and Feng, Haiwen and Black, Michael J and Bolkart, Timo},
  journal={ACM Transactions on Graphics (ToG)},
  volume={40},
  number={4},
  pages={1--13},
  year={2021},
  publisher={ACM New York, NY, USA}
}

@inproceedings{wang20243d,
  title={3d face reconstruction with the geometric guidance of facial part segmentation},
  author={Wang, Zidu and Zhu, Xiangyu and Zhang, Tianshuo and Wang, Baiqin and Lei, Zhen},
  booktitle={Proceedings of the IEEE/CVF Conference on Computer Vision and Pattern Recognition},
  pages={1672--1682},
  year={2024}
}

@inproceedings{liu2025lucas,
  title={LUCAS: Layered Universal Codec Avatars},
  author={Liu, Di and Deng, Teng and Nam, Giljoo and Rong, Yu and Pidhorskyi, Stanislav and Li, Junxuan and Saragih, Jason and Metaxas, Dimitris N and Cao, Chen},
  booktitle={Proceedings of the Computer Vision and Pattern Recognition Conference},
  pages={21127--21137},
  year={2025}
}

@inproceedings{ma20243d,
  title={3d gaussian blendshapes for head avatar animation},
  author={Ma, Shengjie and Weng, Yanlin and Shao, Tianjia and Zhou, Kun},
  booktitle={ACM SIGGRAPH 2024 Conference Papers},
  pages={1--10},
  year={2024}
}

@inproceedings{rota2024revising,
  title={Revising densification in gaussian splatting},
  author={Rota Bul{\`o}, Samuel and Porzi, Lorenzo and Kontschieder, Peter},
  booktitle={European Conference on Computer Vision},
  pages={347--362},
  year={2024},
  organization={Springer}
}

@article{kheradmand20243d,
  title={3d gaussian splatting as markov chain monte carlo},
  author={Kheradmand, Shakiba and Rebain, Daniel and Sharma, Gopal and Sun, Weiwei and Tseng, Yang-Che and Isack, Hossam and Kar, Abhishek and Tagliasacchi, Andrea and Yi, Kwang Moo},
  journal={Advances in Neural Information Processing Systems},
  volume={37},
  pages={80965--80986},
  year={2024}
}

@inproceedings{zhang2024pixel,
  title={Pixel-gs: Density control with pixel-aware gradient for 3d gaussian splatting},
  author={Zhang, Zheng and Hu, Wenbo and Lao, Yixing and He, Tong and Zhao, Hengshuang},
  booktitle={European Conference on Computer Vision},
  pages={326--342},
  year={2024},
  organization={Springer}
}

@article{deng2024efficient,
	title={Efficient Density Control for 3D Gaussian Splatting},
	author={Deng, Xiaobin and Diao, Changyu and Li, Min and Yu, Ruohan and Xu, Duanqing},
	journal={arXiv preprint arXiv:2411.10133},
	year={2024}
}

@article{zheng2025gradient,
	author = {Zhou, Zheng and Xiong, Yu-Jie and Xia, Chun-Ming and Zhang, Jia-Chen and Zhan, Hong-Jian},
	year = {2025},
	month = {08},
	title = {Gradient-Direction-Aware Density Control for 3D Gaussian Splatting},
	journal={arXiv preprint arXiv:2508.09239},
}

@inproceedings{tang2025gaf,
  title={Gaf: Gaussian avatar reconstruction from monocular videos via multi-view diffusion},
  author={Tang, Jiapeng and Davoli, Davide and Kirschstein, Tobias and Schoneveld, Liam and Niessner, Matthias},
  booktitle={Proceedings of the Computer Vision and Pattern Recognition Conference},
  pages={5546--5558},
  year={2025}
}

@inproceedings{saunders2025gasp,
  title={GASP: Gaussian Avatars with Synthetic Priors},
  author={Saunders, Jack and Hewitt, Charlie and Jian, Yanan and Kowalski, Marek and Baltrusaitis, Tadas and Chen, Yiye and Cosker, Darren and Estellers, Virginia and Gyd{\'e}, Nicholas and Namboodiri, Vinay P and others},
  booktitle={Proceedings of the Computer Vision and Pattern Recognition Conference},
  pages={271--280},
  year={2025}
}

@inproceedings{xu2024gaussian,
  title={Gaussian head avatar: Ultra high-fidelity head avatar via dynamic gaussians},
  author={Xu, Yuelang and Chen, Benwang and Li, Zhe and Zhang, Hongwen and Wang, Lizhen and Zheng, Zerong and Liu, Yebin},
  booktitle={Proceedings of the IEEE/CVF conference on computer vision and pattern recognition},
  pages={1931--1941},
  year={2024}
}

@inproceedings{li2025rgbavatar,
  title={RGBAvatar: Reduced Gaussian Blendshapes for Online Modeling of Head Avatars},
  author={Li, Linzhou and Li, Yumeng and Weng, Yanlin and Zheng, Youyi and Zhou, Kun},
  booktitle={Proceedings of the Computer Vision and Pattern Recognition Conference},
  pages={10747--10757},
  year={2025}
}

@article{guo2019relightables,
  title={The relightables: Volumetric performance capture of humans with realistic relighting},
  author={Guo, Kaiwen and Lincoln, Peter and Davidson, Philip and Busch, Jay and Yu, Xueming and Whalen, Matt and Harvey, Geoff and Orts-Escolano, Sergio and Pandey, Rohit and Dourgarian, Jason and others},
  journal={ACM Transactions on Graphics (ToG)},
  volume={38},
  number={6},
  pages={1--19},
  year={2019},
  publisher={ACM New York, NY, USA}
}

@inproceedings{yang2023towards,
  title={Towards practical capture of high-fidelity relightable avatars},
  author={Yang, Haotian and Zheng, Mingwu and Feng, Wanquan and Huang, Haibin and Lai, Yu-Kun and Wan, Pengfei and Wang, Zhongyuan and Ma, Chongyang},
  booktitle={SIGGRAPH Asia 2023 Conference Papers},
  pages={1--11},
  year={2023}
}

@inproceedings{wang2025steepest,
  title={Steepest Descent Density Control for Compact 3D Gaussian Splatting},
  author={Wang, Peihao and Wang, Yuehao and Wang, Dilin and Mohan, Sreyas and Fan, Zhiwen and Wu, Lemeng and Cai, Ruisi and Yeh, Yu-Ying and Wang, Zhangyang and Liu, Qiang and others},
  booktitle={Proceedings of the Computer Vision and Pattern Recognition Conference},
  pages={26663--26672},
  year={2025}
}

@article{egger20203d,
  title={3d morphable face models—past, present, and future},
  author={Egger, Bernhard and Smith, William AP and Tewari, Ayush and Wuhrer, Stefanie and Zollhoefer, Michael and Beeler, Thabo and Bernard, Florian and Bolkart, Timo and Kortylewski, Adam and Romdhani, Sami and others},
  journal={ACM Transactions on Graphics (ToG)},
  volume={39},
  number={5},
  pages={1--38},
  year={2020},
  publisher={ACM New York, NY, USA}
}

@inproceedings{saito2024relightable,
  title={Relightable gaussian codec avatars},
  author={Saito, Shunsuke and Schwartz, Gabriel and Simon, Tomas and Li, Junxuan and Nam, Giljoo},
  booktitle={Proceedings of the IEEE/CVF conference on computer vision and pattern recognition},
  pages={130--141},
  year={2024}
}

@inproceedings{zielonka2022towards,
  title={Towards metrical reconstruction of human faces},
  author={Zielonka, Wojciech and Bolkart, Timo and Thies, Justus},
  booktitle={European conference on computer vision},
  pages={250--269},
  year={2022},
  organization={Springer}
}

@inproceedings{zhang2018unreasonable,
  title={The unreasonable effectiveness of deep features as a perceptual metric},
  author={Zhang, Richard and Isola, Phillip and Efros, Alexei A and Shechtman, Eli and Wang, Oliver},
  booktitle={Proceedings of the IEEE conference on computer vision and pattern recognition},
  pages={586--595},
  year={2018}
}

@article{kingma2014adam,
  title={Adam: A method for stochastic optimization},
  author={Kingma, Diederik P and Ba, Jimmy},
  journal={arXiv preprint arXiv:1412.6980},
  year={2014}
}

@inproceedings{lin2022robust,
  title={Robust high-resolution video matting with temporal guidance},
  author={Lin, Shanchuan and Yang, Linjie and Saleemi, Imran and Sengupta, Soumyadip},
  booktitle={Proceedings of the IEEE/CVF Winter Conference on Applications of Computer Vision},
  pages={238--247},
  year={2022}
}

@inproceedings{kim2024color,
  title={Color-cued efficient densification method for 3d gaussian splatting},
  author={Kim, Sieun and Lee, Kyungjin and Lee, Youngki},
  booktitle={Proceedings of the IEEE/CVF Conference on Computer Vision and Pattern Recognition},
  pages={775--783},
  year={2024}
}

@inproceedings{mallick2024taming,
  title={Taming 3dgs: High-quality radiance fields with limited resources},
  author={Mallick, Saswat Subhajyoti and Goel, Rahul and Kerbl, Bernhard and Steinberger, Markus and Carrasco, Francisco Vicente and De La Torre, Fernando},
  booktitle={SIGGRAPH Asia 2024 Conference Papers},
  pages={1--11},
  year={2024}
}

@article{zhou2025perceptual,
  title={Perceptual-GS: Scene-adaptive Perceptual Densification for Gaussian Splatting},
  author={Zhou, Hongbi and Ni, Zhangkai},
  journal={arXiv preprint arXiv:2506.12400},
  year={2025}
}

@inproceedings{zielonka2025gaussian,
  title={Gaussian eigen models for human heads},
  author={Zielonka, Wojciech and Bolkart, Timo and Beeler, Thabo and Thies, Justus},
  booktitle={Proceedings of the Computer Vision and Pattern Recognition Conference},
  pages={15930--15940},
  year={2025}
}

@inproceedings{zheng2022avatar,
  title={Im avatar: Implicit morphable head avatars from videos},
  author={Zheng, Yufeng and Abrevaya, Victoria Fern{\'a}ndez and B{\"u}hler, Marcel C and Chen, Xu and Black, Michael J and Hilliges, Otmar},
  booktitle={Proceedings of the IEEE/CVF conference on computer vision and pattern recognition},
  pages={13545--13555},
  year={2022}
}

@article{liao2023hhavatar,
  title={HHAvatar: Gaussian Head Avatar with Dynamic Hairs},
  author={Liao, Zhanfeng and Xu, Yuelang and Li, Zhe and Li, Qijing and Zhou, Boyao and Bai, Ruifeng and Xu, Di and Zhang, Hongwen and Liu, Yebin},
  journal={arXiv e-prints},
  pages={arXiv--2312},
  year={2023}
}

@inproceedings{giebenhain2024npga,
  title={Npga: Neural parametric gaussian avatars},
  author={Giebenhain, Simon and Kirschstein, Tobias and R{\"u}nz, Martin and Agapito, Lourdes and Nie{\ss}ner, Matthias},
  booktitle={SIGGRAPH Asia 2024 Conference Papers},
  pages={1--11},
  year={2024}
}

@inproceedings{grassal2022neural,
  title={Neural head avatars from monocular rgb videos},
  author={Grassal, Philip-William and Prinzler, Malte and Leistner, Titus and Rother, Carsten and Nie{\ss}ner, Matthias and Thies, Justus},
  booktitle={Proceedings of the IEEE/CVF conference on computer vision and pattern recognition},
  pages={18653--18664},
  year={2022}
}

@article{wang2025srm,
	title={SRM-Hair: Single Image Head Mesh Reconstruction via 3D Morphable Hair},
	author={Wang, Zidu and Zhao, Jiankuo and Xu, Miao and Zhu, Xiangyu and Lei, Zhen},
	journal={arXiv preprint arXiv:2503.06154},
	year={2025}
}

@article{wang2025pc,
	title={PC-Talk: Precise Facial Animation Control for Audio-Driven Talking Face Generation},
	author={Wang, Baiqin and Zhu, Xiangyu and Shen, Fan and Xu, Hao and Lei, Zhen},
	journal={arXiv preprint arXiv:2503.14295},
	year={2025}
}

@article{van2008tsne,
	author = {van der Maaten, Laurens and Hinton, Geoffrey and Rachmad, Yoesoep},
	year = {2008},
	month = {11},
	pages = {2579-2605},
	title = {Viualizing data using t-SNE},
	volume = {9},
	journal = {Journal of Machine Learning Research}
}

@inproceedings{zheng2022imavatar,
	title={{I} {M} {Avatar}: Implicit Morphable Head Avatars from Videos},
	author={Zheng, Yufeng and Abrevaya, Victoria Fernández and Bühler, Marcel C. and Chen, Xu and Black, Michael J. and Hilliges, Otmar},
	booktitle = {Computer Vision and Pattern Recognition (CVPR)},
	year = {2022}
}

@inproceedings{wang2024s2td,
	title={S2td-face: Reconstruct a detailed 3d face with controllable texture from a single sketch},
	author={Wang, Zidu and Zhu, Xiangyu and Yu, Jiang and Zhang, Tianshuo and Lei, Zhen},
	booktitle={Proceedings of the 32nd ACM International Conference on Multimedia},
	pages={6453--6462},
	year={2024}
}

@article{wang2025bfsm,
	title={BFSM: 3D Bidirectional Face-Skull Morphable Model},
	author={Wang, Zidu and Xu, Meng and Xu, Miao and Ma, Hengyuan and Zhao, Jiankuo and Li, Xutao and Zhu, Xiangyu and Lei, Zhen},
	journal={arXiv preprint arXiv:2509.24577},
	year={2025}
}

@article{chu2024generalizable,
	title={Generalizable and animatable gaussian head avatar},
	author={Chu, Xuangeng and Harada, Tatsuya},
	journal={Advances in Neural Information Processing Systems},
	volume={37},
	pages={57642--57670},
	year={2024}
}

@inproceedings{he2025lam,
	title={LAM: large avatar model for one-shot animatable gaussian head},
	author={He, Yisheng and Gu, Xiaodong and Ye, Xiaodan and Xu, Chao and Zhao, Zhengyi and Dong, Yuan and Yuan, Weihao and Dong, Zilong and Bo, Liefeng},
	booktitle={Proceedings of the Special Interest Group on Computer Graphics and Interactive Techniques Conference Conference Papers},
	pages={1--13},
	year={2025}
}

@inproceedings{zhang2025guava,
	title={Guava: Generalizable upper body 3d gaussian avatar},
	author={Zhang, Dongbin and Liu, Yunfei and Lin, Lijian and Zhu, Ye and Li, Yang and Qin, Minghan and Li, Yu and Wang, Haoqian},
	booktitle={Proceedings of the IEEE/CVF International Conference on Computer Vision},
	pages={14205--14217},
	year={2025}
}
}


\end{document}


\setcounter{page}{1}
\maketitlesupplementary

\section{Overview}
This supplementary material presents additional methodological details, experimental results, and an ethical discussion. Sec.~\ref{sec:method} elaborates on offset activation, UV sampling, silhouette coefficient, and the Summed-Area Table. Sec.~\ref{sec:exp} first provides extended details on the experimental setup for the baseline methods, then visualizes the FTC results to facilitate a more comprehensive understanding, and includes additional ablation studies such as hyperparameter analyses. It also presents further qualitative results, per-identity quantitative evaluations, and a visualization of the learned offset maps for spatial analysis. In addition, a dedicated section discusses the limitations of our optimization-based framework, including comparisons with recent generalizable single-image reconstruction methods. Finally, Sec.~\ref{sec:ethical} discusses data handling practices, potential societal harms, and measures to prevent misuse.

\section{Method}
\label{sec:method}
\subsection{Offset Activation Details.}
To ensure numerical stability and constrain the influence of the predicted offsets 
\(\delta_i = \{\delta_{\mu}, \delta_{s}, \delta_{r}, \delta_{\alpha}, \delta_{c}\}\) 
on the final Gaussian parameters, we apply carefully chosen nonlinear activation functions to each component extracted from the UV feature offset map \(\Delta_{\text{map}}\). These activations are designed to restrict each predicted quantity within a semantically meaningful and physically valid range, as described below:
\begin{itemize}
    \item \textbf{Position offset} \(\delta_{\mu}\): 
     Given the typical spatial extent of Gaussians, we constrain each component of the position offset to the interval \([-0.1, 0.1]\) using a hyperbolic tanh activation:
    \begin{equation}
        \delta_{\mu} = 0.1 \cdot \tanh(x),
    \end{equation}
    where \(x \in \mathbb{R}^3\) is the raw network output.

    \item \textbf{Scaling Offset} \(\delta_{s}\):  
	To ensure the scaling offset remains strictly positive—thus making the multiplication with the coarse scale \(\tilde{s}\) valid—we apply an exponential activation to map the output to the range \((0, +\infty)\):
	\begin{equation}
		\delta_{s} = \exp(x),
	\end{equation}
	where \(x \in \mathbb{R}^3\). The final scale is computed as the element-wise product of \(\delta_s\) and the coarsely deformed scale \(\tilde{s}\), ensuring both positivity and flexible control.

    \item \textbf{Rotation offset} \(\delta_{r}\): 
	The raw rotation offset is represented as an axis-angle vector \(x \in \mathbb{R}^3\). Its magnitude is first normalized to the interval \([-\pi, \pi]\) via:
	\begin{equation}
		\delta_{r}^{\text{angle}} = \pi \cdot \tanh(x),
	\end{equation}
	and then converted to a unit quaternion \(\delta_{r}\) using the standard axis-angle to quaternion transformation. This guarantees a valid rotation that can be composed with the coarsely deformed rotation \(\tilde{r}\) through the Hamilton product.

    \item \textbf{Opacity offset} \(\delta_{\alpha}\): 
	To avoid drastic changes in transparency, the opacity offset is constrained to \([-0.5, 0.5]\):
	\begin{equation}
		\delta_{\alpha} = 0.5 \cdot \tanh(x),
	\end{equation}
	where \(x \in \mathbb{R}\). After adding \(\delta_{\alpha}\) to the coarse opacity \(\tilde{\alpha}\), the final opacity \(\alpha^*\) is further clamped to \([0, 1]\) to ensure physical plausibility.

    \item \textbf{Color offset} \(\delta_{c}\): 
    To limit abrupt color shifts during Gaussian's deformation, we restrict each channel of the color offset to \([-0.7, 0.7]\) via:
    \begin{equation}
        \delta_{c} = 0.7 \cdot \tanh(x),
    \end{equation}
    with \(x \in \mathbb{R}^3\).
\end{itemize}

These bounded activations not only prevent degenerate parameter updates during training (\textit{e.g.}, negative scale or invalid quaternion norms), but also provide a consistent prior for Gaussian parameters, allowing more effective learning from the UV-space offset map \(\Delta_{map}\).

\subsection{UV Sample.}
As shown in Algorithm~\ref{alg:uv_sampling}, we provide a more detailed algorithmic description of UV-Adaptive Sampling. Since certain mesh triangles may not cover any valid pixels after UV rasterization, it becomes infeasible to define the UV coordinates of the associated Gaussian primitives using pixel-based sampling. To address this, we incorporate an analytical fallback strategy that samples barycentric coordinates uniformly within the triangle, enabling the assignment of UV coordinates even when no valid pixels are present.

\begin{figure}[htb]
    \centering
    \includegraphics[width=1\linewidth]{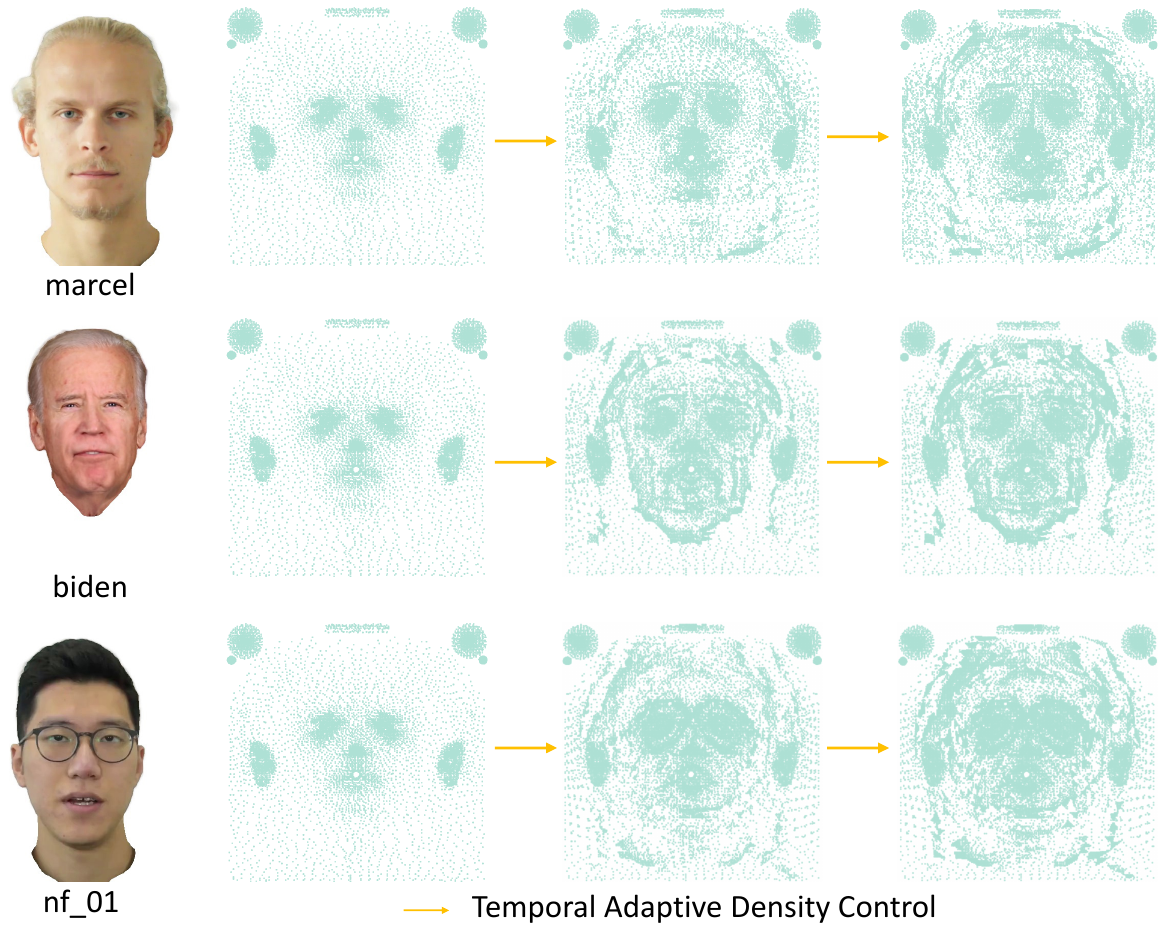}
    \caption{Visualization of UV-Adaptive Sampling's points.}
    \vspace*{-5pt}
    \label{fig:sample}
\end{figure}

As illustrated in Fig.~\ref{fig:sample}, after applying Temporal Adaptive Density Control, our UV-Adaptive Sampling adaptively allocates more Gaussians to regions requiring finer details across different identities. In particular, the central facial region of \textit{biden} contains more points due to the presence of facial wrinkles, which demand higher density. Additionally, the density in \textit{nf\_01}'s eye region significantly increases because \textit{nf\_01} wears glasses, which typically require more Gaussians for accurate modeling.

\begin{algorithm}[!th]
	\caption{UV-Adaptive Sampling}
	\label{alg:uv_sampling}
	\textbf{Input}: \\
	\quad $\mathcal{B} \in \mathbb{N}^N$: face binding index per Gaussian \\
	\quad $\mathcal{V}_{uv} \in \mathbb{R}^{V \times 2}$: UV coordinates of mesh vertices \\
	\quad $\mathcal{F}_{uv} \in \mathbb{N}^{F \times 3}$: triangle face indices over UV mesh \\
	\quad $R$: UV rasterization resolution (e.g., $R=256$) \\
	\textbf{Output}: \\
	\quad $\mathcal{U} \in \mathbb{R}^{N \times 2}$: UV coordinates per Gaussian \\
	\begin{algorithmic}[1]
		\STATE Rasterize UV mesh $(\mathcal{V}_{uv}, \mathcal{F}_{uv})$ into $R \times R$ grid\\
		\quad $\rightarrow (\mathcal{M}_{face}, \mathcal{M}_{bary}) \in \mathbb{Z}^{R \times R}, \mathbb{R}^{R \times R \times 3}$\\
		\STATE Initialize per-face pixel pool map $\mathcal{P}[f] \gets \emptyset,\ \forall f \in \mathcal{F}_{uv}$
		\FOR{each pixel $(i,j)$ where $\mathcal{M}_{face}[i,j] \neq -1$}
		\STATE $f \gets \mathcal{M}_{face}[i,j]$
		\STATE Append $(i,j), \mathcal{M}_{bary}[i,j]$ into $\mathcal{P}[f]$
		\ENDFOR
		\STATE Faces with valid pixels: $\mathcal{F}_{valid} \gets \{f \mid \mathcal{P}[f] \neq \emptyset \}$
		\STATE Initialize barycentric buffer $\mathbf{b} \in \mathbb{R}^{N \times 3}$, face index buffer $\mathbf{f}_{idx} \in \mathbb{N}^{N}$
		\FOR{each face $f$ that appears in bindings $\mathcal{B}$}
		\STATE $\mathcal{I}_f \gets \{ i \mid \mathcal{B}[i] = f \}$ \COMMENT{Indices of points bound to $f$}
		\STATE $C_f \gets |\mathcal{I}_f|$
		\IF{$f \in \mathcal{F}_{valid}$}
		\STATE \textit{Pixel-based Sampling}
		\STATE Let $\mathbf{p}_f$ be sorted pixel list of face $f$ in $\mathcal{P}[f]$
		\IF{$|\mathbf{p}_f| \ge C_f$}
		\STATE Select $C_f$ evenly spaced pixels from $\mathbf{p}_f$
		\ELSE
		\STATE Select all $|\mathbf{p}_f|$ pixels and sample extra with replacement to reach $C_f$
		\ENDIF
		\STATE Extract barycentric coordinates $\mathbf{b}_f$ from $\mathcal{P}[f]$
		\ELSE
		\STATE \textit{Analytical Sampling} 
		\FOR{$i = 1$ to $C_f$}
		\STATE Sample $u,v \sim \mathcal{U}(0,1)$
		\STATE Compute: $b_0 = 1 - \sqrt{u},\ b_1 = \sqrt{u}(1 - v),\ b_2 = \sqrt{u}v$
		\STATE Append $[b_0, b_1, b_2]$ to $\mathbf{b}_f$
		\ENDFOR
		\ENDIF
		\STATE Assign: $\mathbf{b}[\mathcal{I}_f] \gets \mathbf{b}_f$,\quad $\mathbf{f}_{idx}[\mathcal{I}_f] \gets f$
		\ENDFOR
		\STATE $\mathcal{U} \gets \text{BarycentricReweight}(\mathcal{V}_{uv}, \mathcal{F}_{uv}, \mathbf{f}_{idx}, \mathbf{b})$
	\end{algorithmic}
\end{algorithm}

\subsection{Silhouette Coefficient.}
To determine an appropriate number of clusters \(K\) for grouping structurally similar frames based on FLAME~\cite{li2017learning} parameters, we adopt the Silhouette Coefficient as the selection metric. Given a data point \(i\), the silhouette score \(s(i)\) is defined as:
\begin{equation}
    s(i) = \frac{b(i) - a(i)}{\max \left\{ a(i),\ b(i) \right\}}, \quad s(i) \in [-1, 1],
\end{equation}
where \(a(i)\) denotes the mean intra-cluster distance of point \(i\), and \(b(i)\) represents the smallest mean distance between point \(i\) and all points in the nearest neighboring cluster. These quantities are computed as:
\begin{align}
    a(i) &= \frac{1}{|C_i| - 1} \sum_{\substack{j \in C_i \\ j \neq i}} d(i, j), \\
    b(i) &= \min_{k \neq i} \left( \frac{1}{|C_k|} \sum_{j \in C_k} d(i, j) \right),
\end{align}
where \(C_i\) is the cluster to which point \(i\) belongs, and \(d(i, j)\) denotes the Euclidean distance in the PCA-projected FLAME parameter space. A higher average silhouette score across all points indicates better-defined and more compact clustering. We evaluate the silhouette score across varying \(K \in [5,12]\), and select the value that maximizes the overall score.

\subsection{Summed-Area Table.}
To efficiently compute aggregated error values over rectangular regions in the screen-space image domain, we adopt the two-dimensional \textit{Summed-Area Table (SAT)} strategy, also known as the \textit{integral image}. This data structure allows the summation of any axis-aligned rectangular region in constant time using only four memory accesses.

Given an input error map \(E \in \mathbb{R}^{H \times W}\), where \(H\) and \(W\) denote the image height and width, respectively, the SAT \(I \in \mathbb{R}^{(H+1) \times (W+1)}\) is constructed as:
\begin{equation}
I(u, v) = \sum_{y=0}^{u-1} \sum_{x=0}^{v-1} E(y, x),
\end{equation}
with the boundary conditions \(I(0, v) = I(u, 0) = 0\) for all \(u, v\), which provide zero-padding to simplify index handling. The SAT is built in a single pass using two cumulative summations:
\begin{enumerate}
    \item Compute the row-wise prefix sum for each row.
    \item Compute the column-wise prefix sum of the row-wise sums.
\end{enumerate}
This process is equivalent to:
\begin{equation}
\begin{split}
I(u+1, v+1) = E(u, v) + I(u, v+1) \\ + I(u+1, v) - I(u, v)
\end{split}
\end{equation}

\begin{figure}[ht]
    \centering
    \includegraphics[width=0.95\linewidth]{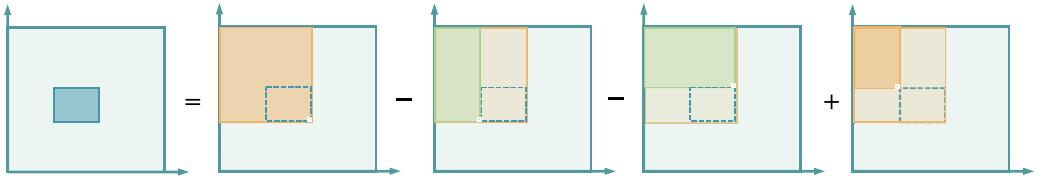}
    \caption{Inclusion–exclusion principle}
    \vspace*{-5pt}
    \label{fig:sat}
\end{figure}

Once the SAT is constructed, the sum over any rectangular region \([x_1, x_2] \times [y_1, y_2]\) can be computed in constant time using:
\begin{equation}
\begin{split}
\sum_{x=x_1}^{x_2} \sum_{y=y_1}^{y_2} E(y,x)
=\, & I(y_2{+}1, x_2{+}1) - I(y_1, x_2{+}1) \\
    & - I(y_2{+}1, x_1) + I(y_1, x_1).
\end{split}
\end{equation}
This formula leverages the inclusion–exclusion principle to compute the sum via four corner indices, as illustrated in Fig.~\ref{fig:sat}. The SAT reduces the computational complexity from \(\mathcal{O}(R^2)\) per Gaussian to \(\mathcal{O}(1)\) per region, enabling efficient error accumulation for tens or hundreds of thousands of Gaussians in real-time optimization.

\section{Experiments}
\label{sec:exp}

\subsection{Baselines}
\label{sec:baseline}
As stated in the main paper, the number of training epochs is set to 10 for GA~\cite{qian2023gaussianavatars}, RGBA~\cite{li2025rgbavatar} and Fate~\cite{zhang2025fate}, 20 for FA~\cite{xiang2024flashavatar}, 30 for SA~\cite{SplattingAvatar:CVPR2024} and 100 for MGA~\cite{chen2024monogaussianavatar} to ensure convergence. Except for the PointAvatar~\cite{Zheng2023pointavatar} dataset, we follow Fate’s training schedule by training all baseline methods for 50 epochs and our method, STAvatar, for 30 epochs. Additionally, following the DECA-based~\cite{feng2021learning} preprocessing pipeline, we further refine the FLAME coefficients for 50 epochs using a learning rate of \(5 \times 10^{-4}\) during testing, which is consistent with the IMAvatar~\cite{zheng2022imavatar} protocol.

\begin{figure}[ht]
	\centering
	\includegraphics[width=1\linewidth]{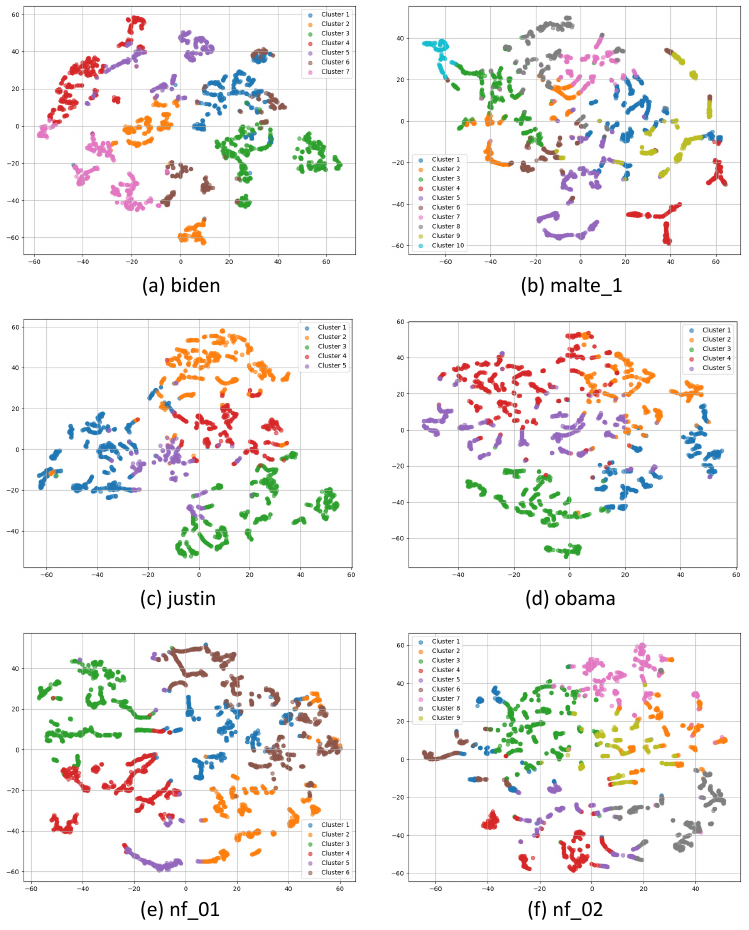}
	\caption{t-SNE visualization of different identities after applying FTC. Each color represents a distinct video frame cluster.}
	\vspace*{-12pt}
	\label{fig:tsne}
\end{figure}

\subsection{t-SNE of FTC. }
To intuitively demonstrate the effectiveness of our FTC strategy, we visualize the clustering results using t-SNE~\cite{van2008tsne} in Fig.~\ref{fig:tsne}. The number of clusters \(K\) is adaptively determined using the silhouette coefficient, which allows FTC to dynamically adjust to the inherent data distribution across identities. As shown in the Fig.~\ref{fig:tsne}, embeddings belonging to different identities are clearly separated in the latent space, suggesting that FTC effectively groups structurally similar frames into coherent clusters, which facilitates the computation of the average densification criterion.

\begin{table}[ht]
	\centering
	\caption{Hyperparameter ablation results. The row in \textbf{bold} indicates the configuration used in the main paper: FPE's \(\mathcal{L}_{\mathrm{d\text{-}ssim}}\) weight $\lambda_1=0.2$, FTC's randomly training epochs $M=1$, and FTC's cluster weights $(0.3,0.6,0.1)$.}
	\resizebox{1\linewidth}{!}{
		\begin{tabular}{@{} l l c c c @{}}
			\toprule
			\multirow{2}{*}{Hyperparameter} & \multirow{2}{*}{Setting} & \multicolumn{3}{c}{Metrics} \\
			\cmidrule(lr){3-5}
			&  & PSNR↑ & SSIM↑ & LPIPS↓ \\
			\midrule
			
			\multirow{3}{*}{$\lambda_1$} 
			& 0.1 & \cellcolor{c2}30.52 & 0.9577 & 0.0312 \\
			& 0.3 & 30.48 & \cellcolor{c2}0.9579 & \cellcolor{c2}0.0311 \\
			& \textbf{0.2} & \cellcolor{c1}{30.63} & \cellcolor{c1}{0.9587} & \cellcolor{c1}{0.0304} \\
			\midrule
			
			\multirow{3}{*}{$M$} 
			& 2 & \cellcolor{c2}30.53 & \cellcolor{c2}0.9582 & \cellcolor{c2}0.0310 \\
			& 0 & 30.42 & 0.9572 & 0.0311 \\
			& \textbf{1} & \cellcolor{c1}{30.63} & \cellcolor{c1}{0.9587} & \cellcolor{c1}{0.0304} \\
			\midrule
			
			\multirow{3}{*}{cluster weights} 
			& (0.5, 0.4, 0.1) & 30.44 & \cellcolor{c2}0.9584 & 0.0312 \\
			& (0.4, 0.5, 0.1) & \cellcolor{c2}30.50 & 0.9583 & \cellcolor{c2}0.0311 \\
			& \textbf{(0.3, 0.6, 0.1)} & \cellcolor{c1}{30.63} & \cellcolor{c1}{0.9587} & \cellcolor{c1}{0.0304} \\
			\bottomrule
		\end{tabular}
	}
	\vspace*{-12pt}
	\label{tab:hyper_combined}
\end{table}

\subsection{Additional Ablation Study}
We conduct additional ablation studies to further validate the effectiveness of each proposed component and the selection of key hyper-parameters. All results are summarized in Tab.~\ref{tab:hyper_combined}, where we compare multiple candidate settings for the FPE's \(\mathcal{L}_{\mathrm{d\text{-}ssim}}\) weight, the FTC's randomly training epochs $M$, and the FTC's cluster weights. The configuration adopted in the main paper is indicated in bold.

\noindent \textbf{FPE: \(\mathcal{L}_{\mathrm{d\text{-}ssim}}\) weight. }  
We examine three choices for the perceptual weighting coefficient. The setting $\lambda_1=0.2$ consistently achieves the highest PSNR, SSIM, and lowest LPIPS, demonstrating that it provides a more balanced trade-off between pixel-level reconstruction and perceptual fidelity. This result aligns with the intuition that moderate perceptual emphasis yields better alignment with human visual sensitivity.

\noindent \textbf{FTC: Randomly training epochs $M$. }  
We further evaluate the influence of the FTC training strategy by varying the number of random training epochs $M$. The configuration $M=1$ produces the best performance, indicating that performing clustered training followed by a single fine-tuning epoch on the entire training set is sufficient and effective. This observation verifies the necessity and contribution of FTC in facilitating the computation of the densification criterion which benefits the reconstruction of transiently visible regions.

\noindent \textbf{FTC: Cluster weights. }  
Besides, we test different FLAME-based cluster weight combinations to assess their impact on the clustering quality. The best-performing setting $(0.3, 0.6, 0.1)$ (corresponding to expression, pose, and translation components) assigns a larger weight to pose-related components, highlighting that pose cues dominate the temporal variations relevant to our densification criterion. This confirms that emphasizing pose information improves the reliability of cluster formation and benefits the downstream optimization process.

\subsection{Visualization of Offset Maps}
\begin{figure}[h]
	\centering
	\vspace{-5pt}
	\includegraphics[width=1\linewidth]{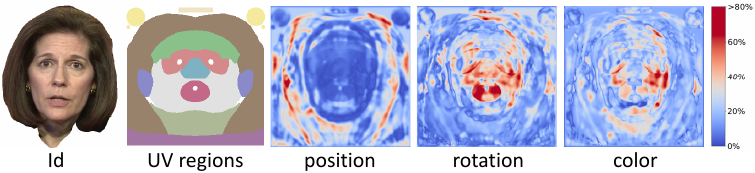}
	\caption{Visualization of learned UV attribute offset maps. Larger magnitudes are primarily observed in high-frequency and highly deformable regions, such as hair and expression-related areas.}
	\vspace{-10pt}
	\label{fig:offset vis}
	\vspace{-3pt}
\end{figure}

To further clarify the spatial effect of the learned offsets, we visualize the distribution of the predicted UV attribute offset maps in Fig.~\ref{fig:offset vis}. The position map is constructed based on the canonical FLAME mesh topology, which does not explicitly model fine-grained structures such as hair. Therefore, it is important to examine whether the learned offsets compensate for these missing details and to identify the regions that benefit most from offset learning.

As illustrated in Fig.~\ref{fig:offset vis}, Gaussians corresponding to hair are primarily associated with the scalp-related UV regions. Notably, the learned offsets exhibit significantly larger magnitudes in high-frequency and highly deformable areas, such as expression-related facial regions (e.g., mouth and eye surroundings) and hair regions. In contrast, offsets remain relatively small in geometrically rigid regions, such as the cheeks and forehead. These observations indicate that the offset learning mechanism adaptively concentrates its modeling capacity on spatially and temporally complex regions, enabling the representation of fine-scale structures (including hair) and dynamic deformations that are not explicitly captured by the underlying FLAME topology.

\subsection{Limitations and Discussion}
\label{sec:limitations}
In recent years, numerous generalizable methods~\cite{chu2024generalizable,he2025lam,zhang2025guava,wang2024s2td,wang2025bfsm} for single-image 3D reconstruction have achieved remarkable performance. Compared with these approaches, our optimization-based framework requires a monocular video sequence and per-subject training, which may limit its applicability in certain scenarios. Although generalizable methods demonstrate strong cross-identity generalization, they are inherently designed for single-image settings and therefore cannot exploit temporal information for subject-specific optimization. As a result, their capacity to learn fine-grained identity-specific characteristics and to capture complex dynamic motions is constrained. Nevertheless, optimization-based methods typically achieve higher identity fidelity due to subject-specific optimization.

\begin{figure}[h]
	\centering
	\vspace{-6pt}
	\includegraphics[width=0.95\linewidth]{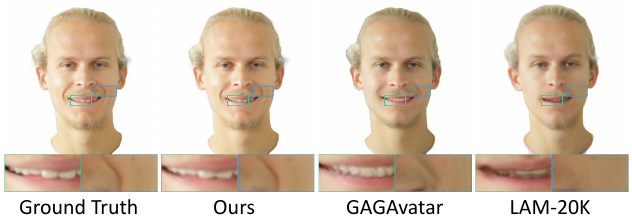}
	\caption{Qualitative comparisons on the INSTA dataset.}
	\label{fig:baselines}
	\vspace{-6pt}
\end{figure}

To illustrate this trade-off, we conduct additional experiments on the INSTA dataset, selecting GAGAvatar and LAM as representative baselines. As reported in Tab.~\ref{tab:baselines} and illustrated in Fig.~\ref{fig:baselines}, these methods struggle to faithfully reconstruct subtle identity details, such as teeth geometry and fine facial wrinkles, whereas our method preserves such high-frequency structures more accurately.

\begin{table}[ht]
	\centering
	\resizebox{0.75\linewidth}{!}{
		\fontsize{9pt}{11pt}\selectfont
		\begin{tabular}{l|ccc}
			\toprule
			Method & PSNR$\uparrow$ & SSIM$\uparrow$ & LPIPS$\downarrow$ \\
			\midrule
			LAM-20K   & 21.71 & 0.8655 & 0.1292 \\
			GAGAvatar & \cellcolor{c2}22.81 & \cellcolor{c2}0.8942 & \cellcolor{c2}0.1042 \\
			Ours      & \cellcolor{c1}{30.63} & \cellcolor{c1}{0.9587} & \cellcolor{c1}{0.0304} \\
			\bottomrule
		\end{tabular}
	}
	\caption{Quantitative comparison with GAGAvatar and LAM on the INSTA dataset.}
	\label{tab:baselines}
	\vspace{-6pt}
\end{table}

\subsection{Self-Reenactment}
\begin{figure*}[ht]
    \centering
    \includegraphics[width=1.0\textwidth]{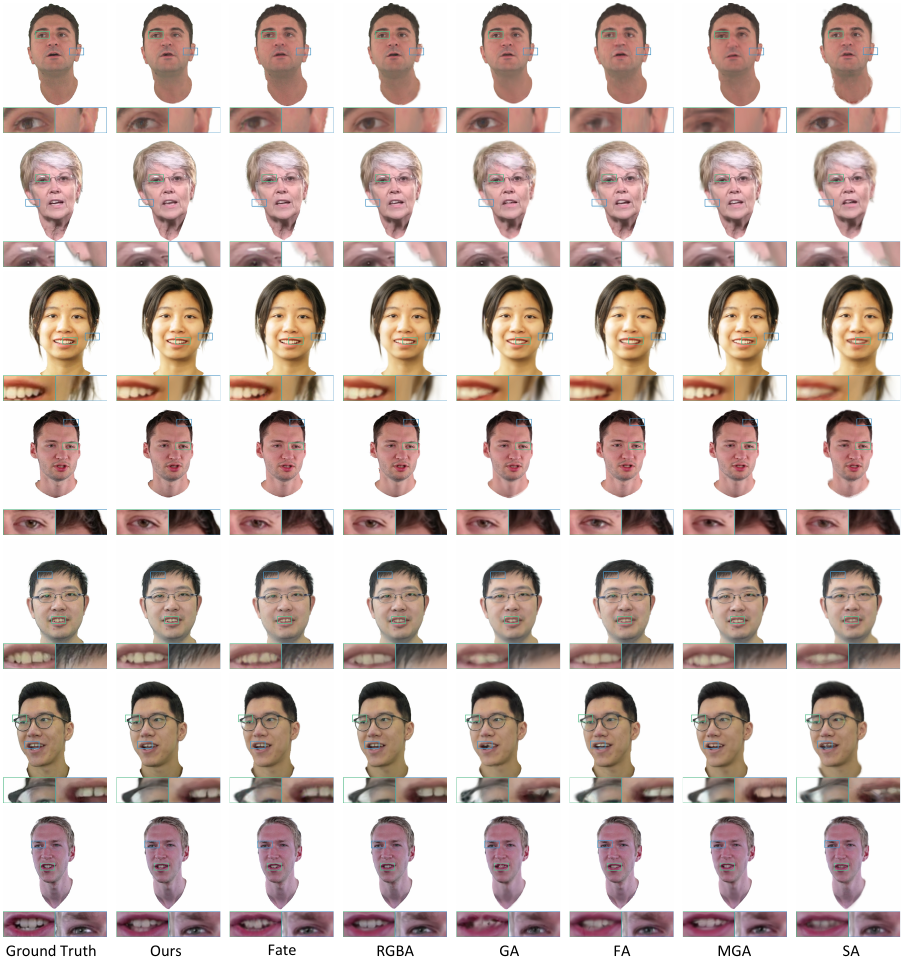}
    \caption{Qualitative comparisons of head reconstruction results across different state-of-the-art methods. Our method STAvatar produces more accurate geometry and fine-grained details, particularly in challenging regions such as occluded inner mouth areas.}
    \vspace*{-5pt}
    \label{fig:app_recon}
\end{figure*}

As shown in Fig.~\ref{fig:app_recon}, STAvatar reconstructs fine-grained geometric and appearance details (e.g., hair strands, eye contours, and glasses frames) more faithfully than prior approaches. With the assistance of Temporal Adaptive Density Control, our model allocates higher density to transient or frequently occluded regions, leading to improved reconstruction of structures such as teeth. As summarized in Tab.~\ref{tab:insta} and Tab.~\ref{tab:other}, although STAvatar does not outperform all baselines on every metric, it achieves the best overall performance on all datasets.

\subsection{Cross-Reenactment}
\begin{figure*}[ht]
    \centering
    \includegraphics[width=0.98\textwidth]{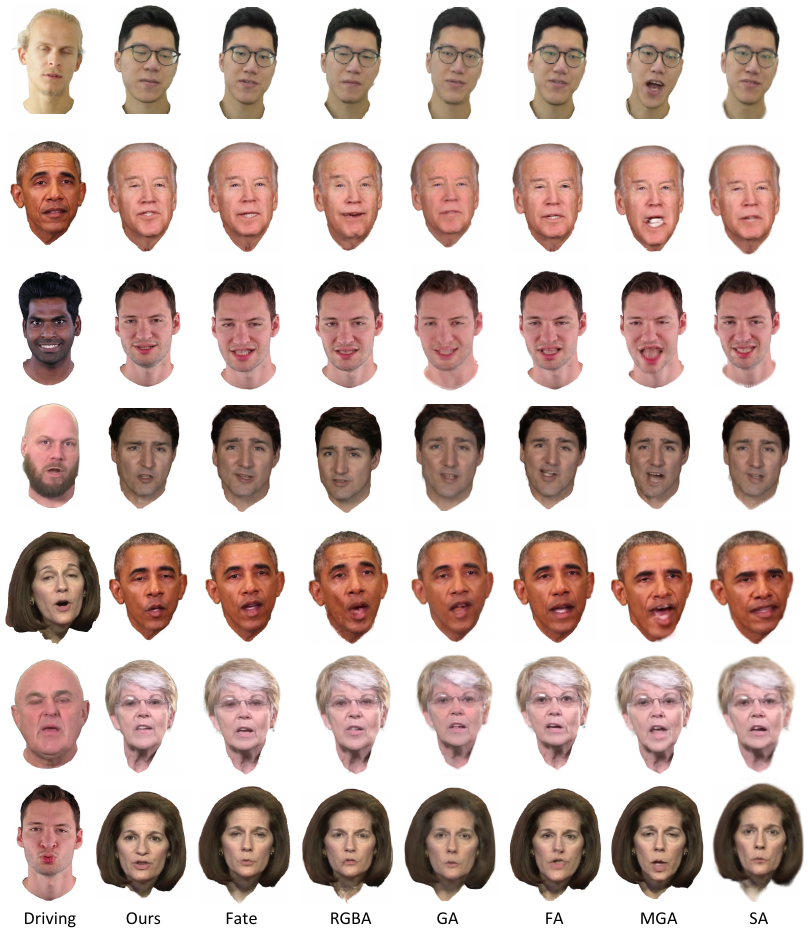}
    \caption{Qualitative comparison of cross-reenactment results with state-of-the-art methods. Our approach achieves achieves faithful expression transfer while maintaining consistent identity-specific geometry and appearance.}
    \vspace*{-12pt}
    \label{fig:app_cross}
\end{figure*}

As illustrated in Fig.~\ref{fig:app_cross}, our method accurately transfers diverse expressions and motions—such as eye-closing, mouth asymmetry, and pouting—from the source actor while preserving the target identity’s geometry and appearance. Even under significant pose variations and complex non-rigid deformations, the reconstructed avatars retain stable identity-related geometry and appearance, demonstrating the robustness of the proposed soft binding and temporal density control.

\section{Ethical Discussion}
\label{sec:ethical}
All datasets used in this study are publicly available and collected in compliance with their respective licenses. While our method enables the generation of realistic and controllable 3D head avatars, we acknowledge its potential misuse in fabricating synthetic content that may infringe on privacy or mislead the public. We explicitly oppose any use of this technology for malicious or unauthorized purposes, including identity fraud or deceptive media creation. Developers and practitioners should act responsibly and follow ethical guidelines to prevent potential harm from misuse.

\begin{table*}[ht]
\centering
\caption{Full comparison of quantitative results with state-of-the-art methods on INSTA dataset.}
\vspace*{-5pt}
\scalebox{0.9}{
\begin{tabular}{ll|cccccccccc}
\toprule
\multicolumn{2}{c}{\multirow{2}{*}{Datasets}} & \multicolumn{10}{c}{INSTA Dataset} \\
\cmidrule(lr){3-12}
\multicolumn{2}{c}{} & bala & biden & justin & malte\_1 & marcel & nf\_01 & nf\_03 & obama & person\_4 & wojtek\_1 \\
\midrule
\multicolumn{1}{l|}{\multirow{7}{*}{PSNR↑}} 
& SplattingAvatar & 27.01 & 29.25 & 28.01 & 27.97 & 29.02 & 24.59 & 27.59 & 23.83 & \cellcolor{c2}{27.19} & 30.35 \\
\multicolumn{1}{l|}{} & MonoGaussianAvatar & 25.39 & 29.38 & 27.32 & 28.25 & 28.27 & 25.09 & 28.01 & 27.10 & 24.48 & 30.16 \\
\multicolumn{1}{l|}{} & GaussianAvatars & 27.41 & 29.11 & 27.49 & 27.69 & 27.84 & 23.71 & 27.27 & 23.40 & 26.04 & 29.89 \\
\multicolumn{1}{l|}{} & FlashAvatar & 28.86 & 28.95 & 27.20 & \cellcolor{c2}{28.46} & 29.07 & 24.90 & 27.87 & 27.24 & 25.80 & 30.61 \\
\multicolumn{1}{l|}{} & RGBAvatar & \cellcolor{c2}{29.11} & \cellcolor{c2}{30.45} & 28.48 & 28.39 & \cellcolor{c2}30.56 & \cellcolor{c2}{26.44} & 27.95 & 27.53 & 24.53 & 30.65 \\
\multicolumn{1}{l|}{} & FateAvatar & 29.02 & 30.28 & \cellcolor{c2}{28.54} & 28.35 & 29.21 & 25.40 & \cellcolor{c2}{28.20} & \cellcolor{c2}{27.93} & 25.40 & \cellcolor{c2}{30.94} \\
\multicolumn{1}{l|}{} & Ours & \cellcolor{c1}{29.26} & \cellcolor{c1}{32.79} & \cellcolor{c1}{29.16} & \cellcolor{c1}{31.73} & \cellcolor{c1}{32.54} & \cellcolor{c1}{28.38} & \cellcolor{c1}{30.72} & \cellcolor{c1}{31.48} & \cellcolor{c1}{29.23} & \cellcolor{c1}{30.98} \\
\hline
\multicolumn{1}{l|}{\multirow{7}{*}{SSIM↑}} 
& SplattingAvatar & 0.9138 & 0.9498 & 0.9570 & 0.9326 & 0.9325 & 0.9225 & 0.9133 & 0.9158 & 0.9452 & 0.9466 \\
\multicolumn{1}{l|}{} & MonoGaussianAvatar & 0.8956 & 0.9546 & 0.9577 & 0.9386 & 0.9322 & 0.9324 & 0.9205 & 0.9426 & 0.9273 & 0.9498 \\
\multicolumn{1}{l|}{} & GaussianAvatars & 0.9287 & 0.9551 & 0.9589 & 0.9356 & 0.9370 & 0.9242 & 0.9179 & 0.9205 & \cellcolor{c2}{0.9488} & 0.9516 \\
\multicolumn{1}{l|}{} & FlashAvatar & 0.9187 & 0.9519 & 0.9513 & 0.9345 & 0.9298 & 0.9307 & 0.9162 & 0.9401 & 0.9377 & 0.9457 \\
\multicolumn{1}{l|}{} & RGBAvatar & \cellcolor{c1}{0.9432} & \cellcolor{c2}{0.9685} & \cellcolor{c2}{0.9705} & \cellcolor{c2}{0.9470} & \cellcolor{c2}0.9452 & \cellcolor{c2}{0.9488} & \cellcolor{c2}{0.9314} & \cellcolor{c2}{0.9542} & 0.9255 & \cellcolor{c1}{0.9591} \\
\multicolumn{1}{l|}{} & FateAvatar & 0.9299 & 0.9646 & 0.9641 & 0.9438 & 0.9403 & 0.9384 & 0.9193 & 0.9511 & 0.9398 & 0.9552 \\
\multicolumn{1}{l|}{} & Ours & \cellcolor{c2}{0.9406} & \cellcolor{c1}{0.9757} & \cellcolor{c1}{0.9706} & \cellcolor{c1}{0.9584} & \cellcolor{c1}{0.9541} & \cellcolor{c1}{0.9524} & \cellcolor{c1}{0.9481} & \cellcolor{c1}{0.9724} & \cellcolor{c1}{0.9575} & \cellcolor{c2}{0.9571} \\
\hline
\multicolumn{1}{l|}{\multirow{7}{*}{LPIPS↓}} 
& SplattingAvatar & 0.1386 & 0.0771 & 0.0811 & 0.0839 & 0.1042 & 0.1325 & 0.1294 & 0.1126 & 0.1101 & 0.0769 \\
\multicolumn{1}{l|}{} & MonoGaussianAvatar & 0.1135 & 0.0486 & 0.0539 & 0.0639 & 0.1338 & 0.1058 & 0.1098 & 0.0581 & 0.1151 & 0.0563 \\
\multicolumn{1}{l|}{} & GaussianAvatars & 0.0976 & 0.0593 & 0.0655 & 0.0714 & 0.1032 & 0.1049 & 0.1138 & 0.0854 & 0.0908 & 0.0593 \\
\multicolumn{1}{l|}{} & FlashAvatar & \cellcolor{c2}{0.0405} & 0.0315 & 0.0384 & \cellcolor{c2}{0.0401} & 0.0775 & 0.0692 & 0.0642 & 0.0407 & 0.1258 & \cellcolor{c2}{0.0347} \\
\multicolumn{1}{l|}{} & RGBAvatar & 0.0535 & \cellcolor{c2}{0.0260} & \cellcolor{c2}{0.0327} & 0.0409 & \cellcolor{c2}0.0655 & \cellcolor{c2}{0.0577} & 0.0666 & \cellcolor{c2}{0.0364} & 0.1029 & 0.0361 \\
\multicolumn{1}{l|}{} & FateAvatar & 0.0534 & 0.0338 & 0.0360 & 0.0412 & 0.0681 & 0.0714 & \cellcolor{c2}{0.0603} & 0.0400 & \cellcolor{c2}{0.0691} & 0.0351 \\
\multicolumn{1}{l|}{} & Ours & \cellcolor{c1}{0.0378} & \cellcolor{c1}{0.0166} & \cellcolor{c1}{0.0249} & \cellcolor{c1}{0.0234} & \cellcolor{c1}{0.0432} & \cellcolor{c1}{0.0406} & \cellcolor{c1}{0.0346} & \cellcolor{c1}{0.0182} & \cellcolor{c1}{0.0386} & \cellcolor{c1}{0.0256} \\
\bottomrule
\end{tabular}
}
\label{tab:insta}
\vspace*{8pt}
\end{table*}

\begin{table*}[ht]
\centering
\caption{Full comparison of quantitative results with state-of-the-art methods on the HDFT dataset, PointAvatar dataset, and NerFace dataset.}
\vspace*{-5pt}
\scalebox{0.75}{
\begin{tabular}{ll|cccccccccccc}
\toprule
\multicolumn{2}{c}{\multirow{2}{*}{Datasets}} & \multicolumn{6}{c}{HDFT Dataset} & \multicolumn{3}{c}{PointAvatar Dataset} & \multicolumn{3}{c}{NerFace Dataset} \\
\cmidrule{3-14}
\multicolumn{2}{c}{} & subject1 & subject2 & subject3 & subject4 & subject5 & subject6 & yufeng & marcel & soubhik & person1 & person2 & person3 \\
\midrule
\multicolumn{1}{l|}{\multirow{7}{*}{PSNR↑}}
& SplattingAvatar & 27.30 & 31.46 & 30.04 & 29.08 & 19.71 & 18.52 & 25.06 & 26.17 & 23.55 & 24.59 & 26.24 & 27.59 \\
\multicolumn{1}{l|}{}
& MonoGaussianAvatar & 28.57 & 32.89 & 30.30 & \cellcolor{c2}30.15 & 20.42 & 19.92 & \cellcolor{c2}28.86 & 25.90 & \cellcolor{c2}28.97 & 25.09 & 26.46 & 28.01 \\
\multicolumn{1}{l|}{}
& GaussianAvatars & 26.21 & 31.06 & 29.63 & 28.69 & 18.16 & 16.74 & 24.93 & 26.15 & 22.77 & 23.71 & 26.25 & 27.27 \\
\multicolumn{1}{l|}{}
& FlashAvatar & 28.88 & 32.77 & 29.25 & 30.13 & 20.37 & 19.60 & 25.86 & 26.04 & 26.67 & 24.90 & \cellcolor{c2}28.10 & 27.87 \\
\multicolumn{1}{l|}{}
& RGBAvatar & 28.83 & 30.30 & 29.57 & 29.44 & \cellcolor{c1}20.97 & \cellcolor{c1}21.23 & 26.34 & \cellcolor{c2}27.35 & 26.32 & \cellcolor{c2}26.44 & 26.99 & 27.95 \\
\multicolumn{1}{l|}{}
& FateAvatar & \cellcolor{c2}28.98 & \cellcolor{c2}33.10 & \cellcolor{c2}30.59 & \cellcolor{c1}30.24 & 20.52 & 19.66 & \cellcolor{c1}29.07 & 26.77 & \cellcolor{c1}29.25 & 25.40 & 27.75 & \cellcolor{c2}28.20 \\
\multicolumn{1}{l|}{}
& Ours & \cellcolor{c1}30.38 & \cellcolor{c1}34.39 & \cellcolor{c1}32.54 & 28.45 & \cellcolor{c2}20.76 & \cellcolor{c2}21.11 & 27.85 & \cellcolor{c1}28.81 & 28.10 & \cellcolor{c1}28.38 & \cellcolor{c1}31.14 & \cellcolor{c1}30.72 \\
\hline
\multicolumn{1}{l|}{\multirow{7}{*}{SSIM↑}}
& SplattingAvatar & 0.8827 & 0.9679 & 0.9556 & 0.9421 & 0.8366 & 0.7684 & 0.8757 & 0.9198 & 0.8858 & 0.9225 & 0.9357 & 0.9133 \\
\multicolumn{1}{l|}{}
& MonoGaussianAvatar & 0.9164 & 0.9745 & 0.9588 & 0.9553 & 0.8743 & 0.8231 & \cellcolor{c1}0.9258 & 0.9154 & \cellcolor{c2}0.9419 & 0.9324 & 0.9469 & 0.9205 \\
\multicolumn{1}{l|}{}
& GaussianAvatars & 0.8826 & 0.9687 & 0.9571 & 0.9447 & 0.8400 & 0.7341 & 0.8904 & \cellcolor{c2}0.9302 & 0.9014 & 0.9242 & 0.9417 & 0.9179 \\
\multicolumn{1}{l|}{}
& FlashAvatar & 0.9198 & 0.9728 & 0.9488 & 0.9538 & 0.8592 & 0.8114 & 0.8870 & 0.9126 & 0.9169 & 0.9307 & \cellcolor{c2}0.9524 & 0.9162 \\
\multicolumn{1}{l|}{}
& RGBAvatar & \cellcolor{c2}0.9347 & 0.9705 & \cellcolor{c2}0.9618 & \cellcolor{c1}0.9598 & \cellcolor{c1}0.8806 & \cellcolor{c1}0.8588 & 0.9020 & 0.9288 & 0.9343 & \cellcolor{c2}0.9488 & 0.9513 & \cellcolor{c2}0.9314 \\
\multicolumn{1}{l|}{}
& FateAvatar & 0.9256 & \cellcolor{c2}0.9747 & 0.9593 & 0.9554 & 0.8700 & 0.8271 & 0.9223 & 0.9254 & 0.9385 & 0.9384 & 0.9515 & 0.9193 \\
\multicolumn{1}{l|}{}
& Ours & \cellcolor{c1}0.9456 & \cellcolor{c1}0.9826 & \cellcolor{c1}0.9719 & \cellcolor{c2}0.9562 & \cellcolor{c2}0.8799 & \cellcolor{c2}0.8522 & \cellcolor{c2}0.9236 & \cellcolor{c1}0.9348 & \cellcolor{c1}0.9426 & \cellcolor{c1}0.9524 & \cellcolor{c1}0.9697 & \cellcolor{c1}0.9481 \\
\hline
\multicolumn{1}{l|}{\multirow{7}{*}{LPIPS↓}}
& SplattingAvatar & 0.1753 & 0.0638 & 0.1167 & 0.1200 & 0.3096 & 0.3070 & 0.1512 & 0.1388 & 0.1534 & 0.1325 & 0.0817 & 0.1294 \\
\multicolumn{1}{l|}{}
& MonoGaussianAvatar & 0.1081 & 0.0471 & 0.0977 & 0.0997 & 0.1776 & 0.2049 & 0.0958 & 0.1552 & 0.0830 & 0.1058 & 0.0545 & 0.1098 \\
\multicolumn{1}{l|}{}
& GaussianAvatars & 0.1502 & 0.0617 & 0.1115 & 0.1237 & 0.2446 & 0.3661 & 0.1397 & 0.1263 & 0.1340 & 0.1049 & 0.0709 & 0.1138 \\
\multicolumn{1}{l|}{}
& FlashAvatar & 0.0639 & \cellcolor{c2}0.0296 & 0.0794 & 0.0578 & \cellcolor{c2}0.1200 & \cellcolor{c2}0.1139 & 0.1100 & 0.1051 & 0.0690 & 0.0692 & \cellcolor{c2}0.0331 & 0.0642 \\
\multicolumn{1}{l|}{}
& RGBAvatar & \cellcolor{c2}0.0595 & 0.0327 & \cellcolor{c2}0.0617 & \cellcolor{c2}0.0503 & 0.1532 & 0.1351 & 0.0872 & \cellcolor{c2}0.0767 & 0.1011 & \cellcolor{c2}0.0577 & 0.0408 & 0.0666 \\
\multicolumn{1}{l|}{}
& FateAvatar & 0.0638 & 0.0320 & 0.0643 & 0.0694 & 0.1389 & 0.1413 & \cellcolor{c2}0.0709 & 0.0960 & \cellcolor{c2}0.0660 & 0.0714 & 0.0356 & \cellcolor{c2}0.0603 \\
\multicolumn{1}{l|}{}
& Ours & \cellcolor{c1}0.0356 & \cellcolor{c1}0.0160 & \cellcolor{c1}0.0275 & \cellcolor{c1}0.0410 & \cellcolor{c1}0.1162 & \cellcolor{c1}0.0893 & \cellcolor{c1}0.0505 & \cellcolor{c1}0.0576 & \cellcolor{c1}0.0404 & \cellcolor{c1}0.0406 & \cellcolor{c1}0.0180 & \cellcolor{c1}0.0346 \\
\bottomrule
\end{tabular}
}
\label{tab:other}
\end{table*}

{
	\small
	\bibliographystyle{ieeenat_fullname}
	\bibliography{supplement}
}